\documentclass[twoside]{article}

\usepackage[preprint]{aistats2027}
\usepackage{xcolor}
\usepackage{hyperref}
\hypersetup{
    colorlinks=true,
    citecolor=blue,   
    linkcolor=blue,   
    urlcolor=blue     
}
\usepackage[nameinlink,capitalize]{cleveref}
\usepackage{enumitem}

\crefname{proposition}{proposition}{propositions}
\Crefname{proposition}{Proposition}{Propositions}

\usepackage[round]{natbib}

\usepackage{amssymb}
\usepackage{nameref}
\usepackage{todonotes}
\usepackage{enumitem}

\usepackage{amsmath,amssymb,amsthm}
\usepackage{algorithm}
\usepackage{algpseudocode}

\usepackage{xcolor} 

\newtheorem{lemma}{Lemma}
\newtheorem{Proposition}{Proposition}
\newtheorem{Corollary}{Corollary}

\newcommand{\vzero}{\mathbf{0}}
\newcommand{\allone}{\mathbf{1}}
\newcommand{\nallone}{\allone}
\newcommand{\mzeros}{\mathbf{0}}

\newcommand{\downbetter}{\textcolor{gray}{\scriptsize$\downarrow$}}
\newcommand{\upbetter}{\textcolor{gray}{\scriptsize$\uparrow$}}

\begin{document}

%
\runningtitle{Constraint-Preserving Flow Matching for Spectral Graph Generation}

%

\twocolumn[

\aistatstitle{Constraint-Preserving Flow Matching for\\Spectral Graph Generation}

\aistatsauthor{ Anita Yang \And Ken-ichi Kawarabayashi \And  Yohan Jung }

\aistatsaddress{ The University of Tokyo \And  National Institute of Informatics\\The University of Tokyo \And Jeonbuk National University } ]

\begin{abstract}
Spectral graph generation models graphs through the eigenvalues and eigenvectors of their Laplacians, providing a compact and continuous alternative to direct graph construction. However, these eigensystems must retain the structural properties of graph Laplacians.
Existing approaches preserve eigenvector orthonormality but do not enforce the Laplacian nullspace constraint, potentially producing eigensystems inconsistent with Laplacian structure. To address this, we introduce \textit{Laplacian-Constrained Stiefel Flow Matching} (LaSFM), which preserves the nullspace condition alongside eigenvector orthonormality and positive eigenvalues during generation. We characterize the underlying constrained Stiefel manifold and formulate flow matching on it. We further extend LaSFM to property-guided generation, steering samples toward desired graph properties while maintaining these constraints. Experiments demonstrate improved performance on both spectral and graph-level metrics, along with better alignment with target properties.

\end{abstract}

\section{Introduction}

Graph generation is central to molecular design, drug discovery,
and modeling complex networks
\citep{jin2018junction,lee2023exploring,you2018graphrnn}. However, directly generating graph is challenging because of its discrete and combinatorial nature~\citep{vignac2023digress}.
Spectral graph generation addresses this difficulty by representing
a graph through the eigenvector matrix $U$ and eigenvalue matrix
$\Lambda$ of its Laplacian $L$, rather than generating edges
directly
\citep{martinkus2022spectre,minello2025generating}.
This provides a compact, continuous representation of global graph
structure and enables smooth generative dynamics in spectral space.

\begin{figure}[t]
    \centering
    \includegraphics[width=\linewidth,height=4.6cm]
    {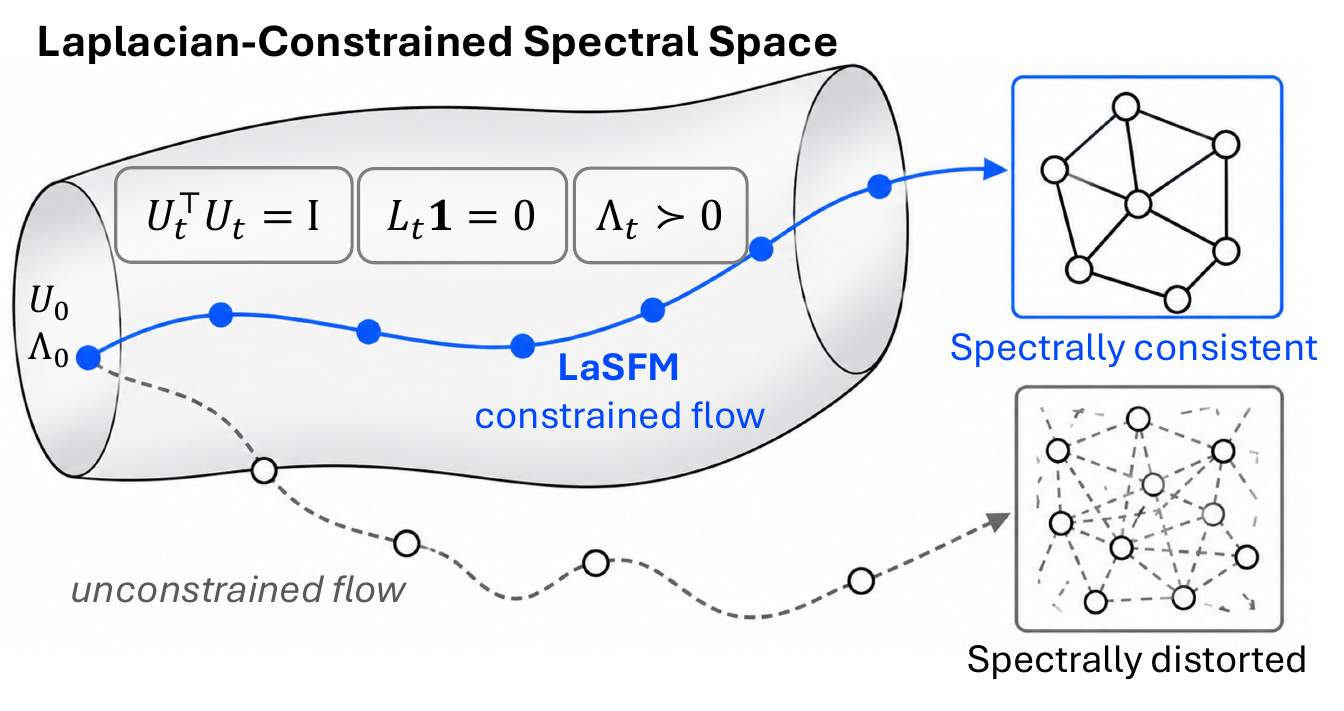}
    \vspace{-5mm}
    \caption{
    Spectral generation can produce distorted structure when eigenvector orthonormality, the Laplacian nullspace constraint, and positive nonzero eigenvalues are not preserved. LaSFM enforces these constraints throughout generation.
    }
    \label{fig:fig1}
\end{figure}

This continuous representation, however, introduces an important
challenge: the generated eigensystem must remain consistent with
the structure of a graph Laplacian. Existing approaches preserve
eigenvector orthonormality, $U^\top U=I$
\citep{martinkus2022spectre,huang2025graph}, but do not explicitly enforce two additional Laplacian properties: the constant-vector nullspace condition, $L\mathbf{1}=0$, and positivity of the nonzero eigenvalues, $\Lambda\succ0$ conditions.
Without preserving these, the Laplacian reconstructed
from the generated eigensystem can have nonzero row sums or lose
positive semidefiniteness, resulting in a matrix that may no longer correspond to a graph Laplacian.

The Laplacian nullspace and eigenvalue positivity constraints directly affect graph generation. If the reconstructed Laplacian does not satisfy these constraints, the subsequent conversion to a discrete graph can alter the generated spectral structure. As a result, structural characteristics encoded in the eigensystem may be lost or distorted in the final graph, undermining the benefits of spectral generation, as illustrated in Fig.~\ref{fig:fig1}.
To avoid this mismatch, we seek to preserve
eigenvector orthonormality, the Laplacian nullspace, and positive
nonzero eigenvalues throughout generation, improving consistency
between the generated eigensystem and the resulting graph.

In this work, we introduce
\textit{Laplacian-Constrained Stiefel Flow Matching} (LaSFM),
a flow-matching framework that generates eigensystems while
preserving the required spectral constraints. We further extend
LaSFM to property-guided generation, allowing samples to be steered
toward desired properties without violating these constraints.
Empirically, LaSFM improves spectral matching, reconstructed graph
quality, and alignment with target properties, while ablations
demonstrate the importance of each constraint.

Our contributions are summarized as follows:
\begin{enumerate}
    \item We characterize the nullspace-constrained Stiefel
    manifold and derive constraint-preserving geodesics and
vector fields.

    \item We develop LaSFM, which combines flow matching on this constrained manifold with a positive
eigenvalue parameterization, preserving orthonormality, the
Laplacian nullspace, and positive nonzero eigenvalues throughout
generation.

    \item We extend LaSFM to gradient-based property guidance and
establish that the spectral constraints remain preserved
under guidance.
\end{enumerate}

\section{Preliminaries}
\label{sec:preliminaries}

\paragraph{Spectral representation of graph Laplacians.}
For an undirected graph with nonnegative adjacency
matrix $A$, the combinatorial Laplacian is $L=D-A$,
where $D_{ii}=\sum_j A_{ij}$.
Since $L$ is symmetric and positive semidefinite,
it admits the eigendecomposition
$L=U\Lambda U^\top$, where $U$ contains
orthonormal eigenvectors and $\Lambda$ contains
nonnegative eigenvalues.

Spectral graph generation typically retains only
$k<n$ eigenpairs to obtain a compact approximation of the graph spectrum:
\begin{equation}
L \approx U_k \Lambda_k U_k^\top,
\end{equation}
where $U_k\in\mathbb{R}^{n\times k}$ contains the retained
eigenvectors and $\Lambda_k\in\mathbb{R}^{k\times k}$ is the
corresponding diagonal eigenvalue matrix.
Hereafter, we omit the subscript $k$ for brevity and consider only
the nonzero eigenvalues and their corresponding eigenvectors.

\paragraph{Structural constraints of Laplacian eigensystems.}
A combinatorial graph Laplacian is known to satisfy four properties~\citep{chung1997spectral}:
\vspace{-2mm}
\begin{enumerate}[label=\textit{(\roman*)}]
    \item Symmetry: $L=L^\top$ \vspace{-1mm}
    \item Positive semidefiniteness: $L\succeq0$ \vspace{-1mm}
    \item Constant-vector nullspace condition: 
    $L\mathbf{1}=0$ \vspace{-1mm}
    \item Nonpositive off-diagonal entries: 
    $L_{ij}\leq0$ for $i\neq j$
\end{enumerate}

For the truncated eigensystem $(U,\Lambda)$, the first three
properties correspond to spectral constraints on the retained
eigenpairs. Symmetry gives an orthonormal eigenbasis, requiring
$U^\top U=I_k$ in the spectral decomposition; positive
semidefiniteness requires the retained nonzero eigenvalues to be
positive, equivalently $\Lambda\succ0$; and the constant-vector
nullspace requires the retained eigenvectors to be orthogonal to
$\mathbf{1}$. We therefore impose the following spectral constraints:
\begin{equation}\label{eq:spectral_constraints}
    U^\top U=I_k,\qquad
    \Lambda\succ0,\qquad
    U^\top\mathbf{1}=\mathbf{0}.
\end{equation}
We do not directly enforce the nonpositive off-diagonal condition
in (iv), as it is generally not preserved under low-rank spectral
truncation.

\paragraph{Spectral graph generation via flow matching.}
Flow matching~\citep{lipman2022flow,albergo2022building,liu2023flow}
learns a time-dependent vector field that transports samples from a
base distribution to the data distribution. In spectral graph
generation, flow matching can be used to generate the eigenvectors
$U$ and eigenvalues $\Lambda$ of graph Laplacians
\citep{huang2025graph}. Specifically, separate vector fields
$V_U$ and $V_\Lambda$ are learned for the eigenvectors and
eigenvalues, and the eigensystem $(U,\Lambda)$ is generated by
integrating these fields from the base distribution
(see Appendix~\ref{app:spectral-flow} for details).

\paragraph{Stiefel flow matching.}
The orthonormality constraint on the eigenvector matrix $U$ naturally motivates modeling its dynamics on the Stiefel manifold, where orthonormality is preserved. Flow matching can also be formulated on this manifold through Stiefel flow matching~\citep{cheng2025stiefel}, which builds on the broader Riemannian framework~\citep{chen2024flow}.

For time index $t\in[0,1]$, let $U_t$ denote the eigenvector matrix
at time $t$, where $U_0$ is sampled from a base distribution on the
Stiefel manifold and $U_1$ is the eigenvector matrix of a graph
Laplacian from the data distribution. To preserve eigenvector
orthonormality throughout the flow, $U_t$ must remain on the Stiefel
manifold
\begin{equation}
    \mathrm{St}(n,k)
    =
    \left\{
        U\in\mathbb{R}^{n\times k}:
        U^\top U=I_k
    \right\}.
\end{equation}
Consequently, its velocity
$\dot U_t:=\frac{d}{dt}U_t$ must lie in the corresponding tangent space
\begin{equation}
    T_{U_t}\mathrm{St}(n,k)
    =
    \left\{
        \Delta\in\mathbb{R}^{n\times k}:
        U_t^\top\Delta+\Delta^\top U_t = \mathbf{0}
    \right\},
    \label{eq:tangent_space}
\end{equation}
obtained by differentiating the orthonormality condition
$U_t^\top U_t=I_k$, which gives
$\dot U_t^\top U_t+U_t^\top\dot U_t=\mathbf{0}$.

To generate eigenvectors on this manifold, Stiefel flow matching learns a tangent vector field $V_U(t,U_t)$ that approximates the geodesic velocity $\dot U_t$
by minimizing
\begin{equation*}
    \mathcal{L}_U
    =
    \mathbb{E}_{t,U_0,U_1}
    \left[
        \left\|
            V_U(t,U_t)-\dot U_t
        \right\|_{U_t}^2
    \right],
    \label{eq:stiefel_fm_loss}
\end{equation*}
where $\|\cdot\|_{U_t}$ denotes the norm induced by
the Stiefel metric. Here, the geodesic velocity $\dot U_t$
is given by
\begin{equation}
    \dot U_t =
    \frac{1}{1-t}\operatorname{Log}_{U_t}(U_1),
    \qquad 0\leq t<1.
    \label{eq:dot_u_t}
\end{equation}
This is obtained by differentiating the geodesic $U_t$
between the source $U_0$ and data $U_1$:
\begin{equation}
    U_t=
    \operatorname{Exp}_{U_0}
    \left(t\operatorname{Log}_{U_0}(U_1)\right).
    \label{eq:geodesic_path}
\end{equation}
Here, $\operatorname{Exp}(\cdot)$ and $\operatorname{Log}(\cdot)$ denote the Riemannian exponential and a logarithm map chosen consistently along the geodesic, respectively ~\citep{cheng2025stiefel}. Further details are provided in Appendix ~\ref{app:stiefel-background}.

\section{Nullspace-Constrained Stiefel Geometry}
\label{sec:geometry}

Stiefel flow matching preserves eigenvector orthonormality in
spectral graph generation~\citep{huang2025graph}, but does not enforce
the constant-vector nullspace condition of graph Laplacians.
We therefore extend the Stiefel dynamics to enforce this additional constraint.
We first construct a geodesic $U_t$ that
satisfies the nullspace condition (Sec. \ref{sec:nullspace_manifold}) and derive its velocity $\dot U_t$ as the
flow-matching target (Sec. \ref{sec:nullspace_geodesic}). We then establish a sufficient condition on the
learned vector field $V_U(t,U_t)$ to maintain it throughout generation (Sec. \ref{sec:nullspace_vector_field}).

\subsection{Nullspace-Constrained Eigenvector Manifold}
\label{sec:nullspace_manifold}

We first establish the relationship between the nullspace condition of the reconstructed Laplacian $L=U\Lambda U^\top$ and the corresponding
eigenvector matrix $U$.

\begin{Proposition}
\label{prop:nullspace}
For a Laplacian matrix $L$, let
    $U\in\operatorname{St}(n,k)$ and let $\Lambda$ be a diagonal matrix
    with positive eigenvalues such that $
    L=U\Lambda U^\top$.    
    Then,
    \vspace{1mm}
    \[
    L\allone =\mzeros
    \quad\text{if and only if}\quad
    U^{\top} \allone=\mzeros.
    \]
\end{Proposition}
\noindent\textit{Proof.}
See Appendix~\ref{app:proofs}.

Proposition~\ref{prop:nullspace} allows us to
enforce the nullspace condition directly on
the retained eigenvector matrix. We therefore define
the nullspace-constrained Stiefel manifold
\begin{equation}
\label{eq:constrained_stiefel}
    \mathcal M_{n,k}
    =
    \left\{
    U\in\mathrm{St}(n,k):
    U^\top\mathbf{1}=\mathbf{0}
    \right\}.
\end{equation}

Since $U^\top\mathbf{1}=\mathbf{0}$, the columns of $U$ lie in the $(n-1)$-dimensional subspace $\mathbf{1}^{\perp}$.
Let $B\in\mathbb R^{n\times(n-1)}$ be an orthonormal basis for this subspace.
By Lemma~\ref{lem:nullspace_reduction} in
Appendix~\ref{app:proofs}, every $U\in\mathcal M_{n,k}$ can be
written as
\begin{equation*}
    U=B\widetilde U,
    \qquad
    \text{for some }\widetilde U\in\mathrm{St}(n-1,k).
\end{equation*}
This provides an alternative approach: perform flow matching on the lower-dimensional Stiefel manifold $\mathrm{St}(n-1,k)$ and map the resulting eigenvectors back through $B$. However, this requires choosing an explicit basis and transforming states, velocities, and guidance gradients between the two coordinate systems. We instead formulate the dynamics directly on $\mathcal M_{n,k}$ in the original node coordinates, avoiding these additional transformations.

\subsection{Nullspace-Preserving Geodesic Interpolation}
\label{sec:nullspace_geodesic}

Stiefel flow matching uses a geodesic path $U_t$ and its velocity
$\dot U_t$ as the flow-matching target. We now establish conditions
under which $U_t$ remains on $\mathcal M_{n,k}$
(Eq.~\ref{eq:constrained_stiefel}), ensuring that the interpolation
preserves the constant-vector nullspace constraint.

\begin{Proposition}
\label{prop:nullspace_geodesic}
Let $U_0,U_1\in\operatorname{St}(n,k)$ satisfy
\begin{equation*}
    U_0^\top\allone=\vzero,
    \qquad
    U_1^\top\allone=\vzero,
\end{equation*}
so that $U_0,U_1\in\mathcal M_{n,k}$.
If $U_0$ and $U_1$ are joined by a unique minimizing Stiefel
geodesic $U_t$, $t\in[0,1]$, then
\begin{equation*}
    U_t^\top\allone=\vzero,
    \qquad
    \forall t\in[0,1].
\end{equation*}
\end{Proposition}

\noindent\textit{Proof.}
See Appendix~\ref{app:proofs}.

Intuitively, Proposition~\ref{prop:nullspace_geodesic} shows that if both endpoint eigenvector matrices satisfy the nullspace constraint, then the minimizing Stiefel geodesic between them preserves $U_t^\top\allone=\vzero$ throughout the training trajectory. Together with Proposition~\ref{prop:nullspace}, this ensures that the reconstructed Laplacian satisfies $L_t\mathbf{1}=\mathbf{0}$ throughout the generation. Additionally, since $U_0$ is randomly sampled from a continuous distribution on the manifold, the minimizing geodesic to $U_1$ is unique almost surely, as cases with multiple shortest paths occur with probability zero~\citep{lee2018riemannian}.

Notably, Proposition~\ref{prop:nullspace_geodesic} requires $U_0$ and $U_1$ to satisfy the nullspace condition. Since the data eigenvectors $U_1$ already satisfy the endpoint conditions, we only need to construct $U_0$ on the Stiefel manifold with $U_0^\top\mathbf{1}=\mathbf{0}$.
We obtain $U_0$ by first sampling a Gaussian noise matrix
$Z\in\mathbb R^{n\times k}$ and projecting it onto
$\mathbf{1}^{\perp}$ using
\begin{equation}
\label{eq:project_centered}
P
=
I_n-\frac{1}{n}\mathbf{1}\mathbf{1}^\top.
\end{equation}
We then orthonormalize the columns of $PZ$ to obtain
$U_0\in\mathcal M_{n,k}$ (see Appendix~\ref{app:training}).

\subsection{Nullspace-Preserving Vector Fields}
\label{sec:nullspace_vector_field}

Next, we derive a sufficient condition on the vector field that
ensures the generated eigenvector matrix remains on
$\mathcal M_{n,k}$ throughout the flow.

\begin{Proposition}
\label{prop:nullspace_velocity}
Let $U_t\in\mathrm{St}(n,k)$ be a
continuously differentiable trajectory
with initial condition
$U_0^\top\mathbf{1}=\mathbf{0}$.
If its velocity satisfies
\begin{equation*}
    \dot U_t^\top\mathbf{1}
    =
    \mathbf{0},
    \qquad
    \forall t\in[0,1],
\end{equation*}
then
\begin{equation*}
    U_t^\top\mathbf{1}
    =
    \mathbf{0},
    \qquad
    \forall t\in[0,1].
\end{equation*}
\end{Proposition}

\noindent\textit{Proof.}
See Appendix~\ref{app:proofs}.

Proposition~\ref{prop:nullspace_velocity} shows that a Stiefel trajectory preserves the nullspace condition if $U_0^\top\mathbf{1}=\mathbf{0}$ and $\dot U_t^\top\mathbf{1}=\mathbf{0}$ for all $t$. Since $U_0$ already satisfies the initial condition by construction (Sec.~\ref{sec:nullspace_geodesic}), we only need to enforce the velocity condition on the learned vector field $V_U(t,U_t)$, which approximates $\dot U_t$.

We implement this through two successive projections of the learned vector field $V_U(t,U_t)$. First, we use $P$ (Eq.~\ref{eq:project_centered}) to project the field onto $\mathbf{1}^{\perp}$, satisfying the nullspace velocity condition. We then apply the Stiefel tangent-space projection~\citep{edelman1998geometry} to preserve eigenvector orthonormality, yielding
\begin{equation}
\label{eq:pproj}
    \widetilde{\Pi}(V_U)
    =
    PV_U
    -
    U_t\operatorname{sym}
    \left(U_t^\top PV_U\right),
\end{equation}
where $V_U:=V_U(t,U_t)$ and $\operatorname{sym}(X):=(X+X^\top)/2$.

We now show that the two-step projection enforces both the Stiefel tangent-space and nullspace velocity conditions required to remain on $\mathcal M_{n,k}$.

\begin{Proposition}
\label{prop:nullspace_projection}
For $U_t\in\mathcal M_{n,k}$
and any model output $V_U(t,U_t)$,
the projected velocity in
Eq.~\ref{eq:pproj} satisfies
\begin{equation*}
\begin{aligned}
    \widetilde{\Pi}(V_U)^\top
    \mathbf{1}
    =\mathbf{0},\quad
    U_t^\top\widetilde{\Pi}(V_U)
    +
    \widetilde{\Pi}(V_U)^\top U_t
    =
    \mathbf{0}.
\end{aligned}
\end{equation*}
Consequently,
$\widetilde{\Pi}(V_U)
\in T_{U_t}\mathcal M_{n,k}$.
\end{Proposition}

\noindent\textit{Proof.}
See Appendix~\ref{app:proofs}.

Proposition~\ref{prop:nullspace_projection} establishes that the projected vector field $\widetilde{\Pi}(V_U)$ lies in the tangent space $T_{U_t}\mathcal M_{n,k}$, meaning that its velocity is constrained to directions along the manifold. Therefore, starting from $\smash{U_0\in\mathcal M_{n,k}}$, the trajectory generated by $\smash{\widetilde{\Pi}(V_U)}$ remains on $\smash{\mathcal M_{n,k}}$, preserving both eigenvector orthonormality and the nullspace condition throughout generation.

\begin{figure}[t]
    \centering
    \includegraphics[width=1.0\linewidth]{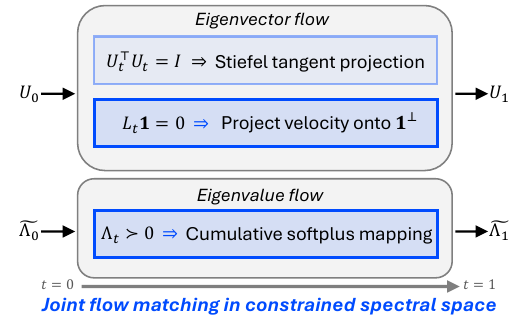}
    \vspace{-5mm}
    \caption{Overview of LaSFM. Beyond eigenvector orthogonality, LaSFM preserves the Laplacian nullspace constraint and eigenvalue positivity throughout generation.}
    \label{fig:overview}
\end{figure}

\begin{algorithm}[t]
\caption{LaSFM Training}
\label{alg:training}
\begin{algorithmic}[1]
\setlength{\abovedisplayskip}{1.5pt}
\setlength{\belowdisplayskip}{1.5pt}
\setlength{\abovedisplayshortskip}{1.5pt}
\setlength{\belowdisplayshortskip}{1.5pt}

\Require Data distribution $p_{\mathrm{data}}$,
LaSFM model $F_\theta$

\State Sample $A_1\sim p_{\mathrm{data}}$ and construct Laplacian $L_1$
\State Compute $(\Lambda_1,U_1)\leftarrow\mathrm{Eig}_k(L_1)$
\State Sample diagonal $\widetilde{\Lambda}_0$ with
$[\widetilde{\Lambda}_0]_{ii}\sim\mathcal N(0,1)$
\State Sample $[U_0]_{ij}\sim \mathcal N(0,1)$
\State Sample $t\sim\mathcal{U}(0,1)$

\vspace{1mm}
\Statex \textbf{Flow construction} \textit{(paths and targets)}

\State Map $\Lambda_1$ to $\widetilde{\Lambda}_1$ using
Eq.~\ref{eq:lambda_map}
\State Project $U_0$ onto $\mathbf{1}^{\perp}$ by Eq.~\ref{eq:project_centered} and orthonormalize 
\State Construct paths
$\smash{(\widetilde{\Lambda}_t,U_t)}$ using Eq.~\ref{eq:lasfm_paths}
\State Construct targets $\smash{(\dot{\widetilde{\Lambda}}_t,\dot U_t)}$ using Eq.~\ref{eq:lasfm_targets}

\vspace{1mm}
\Statex \textbf{Velocity prediction} \textit{(model outputs)}

\State Predict the transformed-eigenvalue and eigenvector velocities
\[
(\widehat V_{\widetilde{\Lambda}_t},\widehat V_{U_t})
\leftarrow
F_\theta(t,U_t,\widetilde{\Lambda}_t)
\]

\State Apply constrained projection
$\widetilde{\Pi}(\widehat V_U)$ from Eq.~\ref{eq:pproj}

\vspace{1mm}
\Statex \textbf{Flow-matching loss} \textit{(training objective)}

\State Compute
\[
\ell_{\widetilde{\Lambda}}
=
\left\|
\widehat V_{\widetilde{\Lambda}_t}
-
\dot{\widetilde{\Lambda}}_t
\right\|_F^2,
\qquad
\ell_U
=
\left\|
\widetilde{\Pi}(\widehat V_{U_t})
-
\dot U_t
\right\|_{U_t}^2
\]

\vspace{1mm}
\Statex \textbf{Parameter update}

\State Update $\theta$ to minimize
$\ell_{\widetilde{\Lambda}}+\ell_U$

\end{algorithmic}
\end{algorithm}

\section{Laplacian-Constrained Stiefel Flow Matching}
\label{sec:lasfm}

Building on the theoretical results developed in the previous section, we introduce \emph{Laplacian-Constrained Stiefel Flow Matching} (LaSFM), a joint flow-matching framework for generating Laplacian eigenvectors and eigenvalues while preserving their spectral constraints. We first introduce the positive eigenvalue parameterization in LaSFM. We then describe the joint learning of eigenvalues and eigenvectors, where the eigenvectors follow nullspace-constrained Stiefel dynamics. An overview is illustrated in~\cref{fig:overview}.

\paragraph{Positive Eigenvalue Parameterization.}
To ensure positive generated eigenvalues, we perform flow matching
over unconstrained diagonal matrices $\widetilde{\Lambda}_t\in\mathbb{R}^{k\times k}$. Then,
the constrained eigenvalue matrix
$\Lambda_t\in\mathbb{R}^{k\times k}$ is obtained through a
cumulative softplus transformation of its diagonal entries:
\begin{equation}
\label{eq:lambda_map}
    [\Lambda_t]_{ii}
    =
    \sum_{j=1}^{i}
    \operatorname{softplus}
    \!\left([\widetilde{\Lambda}_t]_{jj}\right),
    \qquad i=1,\ldots,k,
\end{equation}
where $[\Lambda_t]_{ij}$ denotes the $(i,j)$-th entry of $\Lambda_t$, and $[\Lambda_t]_{ij}=0$ for $i\neq j$.
This ensures $\Lambda_t\succ0$ with
strictly increasing diagonal entries, without constraining the
eigenvalue vector field directly.

\paragraph{Joint Flow Matching.}
We now combine the constrained eigenvector dynamics and positive eigenvalue parameterization into a joint flow-matching formulation. Both vector fields are conditioned on the evolving spectral state $(U_t,\widetilde{\Lambda}_t)$, allowing the eigenvectors and eigenvalues to be jointly modeled.

For the eigenvector component, we sample an initial $U_0\in\mathcal M_{n,k}$ by drawing a Gaussian matrix, projecting
it onto $\mathbf{1}^{\perp}$ using
Eq.~\ref{eq:project_centered}, and orthonormalizing the result,
as described in Sec.~\ref{sec:nullspace_geodesic}. For the eigenvalues,
we sample an initial diagonal matrix
$\widetilde{\Lambda}_0\in\mathbb R^{k\times k}$ with
\[
[\widetilde{\Lambda}_0]_{ii}
\sim\mathcal N(0,1),
\qquad i=1,\ldots,k,
\]
and zero off-diagonal entries. 
The target $\widetilde{\Lambda}_1$ is obtained by inverting Eq.~\ref{eq:lambda_map}, with eigenvalue gaps lower bounded by $\epsilon_\lambda$ before applying the inverse softplus.

Then, we construct the
eigenvalue path $\widetilde \Lambda_t$ by linear interpolation and the
eigenvector path $U_t$  along the nullspace-preserving
minimizing Stiefel geodesic:
\begin{equation}
\label{eq:lasfm_paths}
\widetilde{\Lambda}_t
=
(1-t)\widetilde{\Lambda}_0
+t\widetilde{\Lambda}_1,
\quad
U_t=
\operatorname{Exp}_{U_0}
\left(t\operatorname{Log}_{U_0}(U_1)\right),
\end{equation}
where $U_t$ is the geodesic defined in
Eq.~\ref{eq:geodesic_path}. The corresponding target
velocities are
\begin{equation}
    \label{eq:lasfm_targets}
\dot{\widetilde{\Lambda}}_t
=
\widetilde{\Lambda}_1-\widetilde{\Lambda}_0, \qquad
\dot U_t =
    \frac{1}{1-t}\operatorname{Log}_{U_t}(U_1)
\end{equation}
where $\dot U_t$ is the geodesic velocity defined in Eq.~\ref{eq:dot_u_t}.

Using the target velocities in Eq.~\ref{eq:lasfm_targets}, we jointly train the eigenvalue vector field $V_{\widetilde{\Lambda}}$ and eigenvector vector fields $V_U$ by minimizing
\begin{equation}
\label{eq:lasfm_loss}
\mathcal L_{\mathrm{LaSFM}}
=
\mathbb E\Big[
\underbrace{
\left\|
V_{\widetilde{\Lambda}_t}
-\dot{\widetilde{\Lambda}}_t
\right\|_F^2
}_{\ell_{\widetilde{\Lambda}}}
+
\underbrace{
\left\|
\widetilde{\Pi}(V_{U_t})
-\dot U_t
\right\|_{U_t}^2
}_{\ell_U}
\Big],
\end{equation}
where $\ell_{\widetilde{\Lambda}}$ and $\ell_U$
denote the transformed-eigenvalue and constrained-eigenvector
per-sample flow-matching losses, respectively. Both vector fields are
conditioned on $(t,U_t,\widetilde{\Lambda}_t)$,
$V_{\widetilde{\Lambda}}$ is diagonal, and
$\widetilde{\Pi}$ denotes the constrained tangent-space
projection defined in Eq.~\ref{eq:pproj}.

\paragraph{Architecture and Generation.}
Following \citet{minello2025generating},
we employ bidirectional cross-attention
to jointly model the two spectral components.
Architectural details are provided
in Appendix \ref{app:model}.

During generation, we jointly integrate the learned eigenvector
and eigenvalue velocity fields:
\begin{equation*}
U_t = U_0 + \int_0^t
\widetilde{\Pi}\bigl(\widehat V_{U_s}\bigr)\,ds,\qquad
\widetilde{\Lambda}_t
= \widetilde{\Lambda}_0
+ \int_0^t
\widehat V_{\widetilde{\Lambda}_s}\,ds.
\end{equation*}
The eigenvalues $\Lambda_t$ are then recovered from
$\widetilde{\Lambda}_t$ using Eq.~\ref{eq:lambda_map},
ensuring positivity, while the projected eigenvector
velocity preserves orthonormality and the nullspace
condition. Consequently, the reconstructed matrix
$L_t=U_t\Lambda_t U_t^\top$ remains symmetric,
positive semidefinite, and has zero row sums
throughout the continuous generation process.

\section{Constraint-Preserving Property Guidance}
\label{sec:guidance}

In many applications, we want generated graphs to satisfy certain desired properties. For example, molecular graph generation may target specific electronic properties. Property guidance can steer generation toward these targets. However, directly adding guidance gradients to the spectral dynamics may violate the spectral constraints, causing the trajectory to leave the constrained space. To address this, we extend LaSFM to property-guided generation while preserving the spectral constraints throughout the guided trajectory.

Following gradient-based guidance~\citep{dhariwal2021diffusion},
we steer generated graphs toward desired properties using a
differentiable objective
\[
\mathcal{L}_{\mathrm{guide}}(U,\Lambda).
\]
We modify the eigenvalue and eigenvector vector fields using its gradients:
\begin{equation}
\label{eq:guided_dynamics}
\begin{aligned}
V_{\widetilde{\Lambda}}^{\mathrm{guide}}
&=
V_{\widetilde{\Lambda}}
-
\alpha\nabla_{\widetilde{\boldsymbol{\Lambda}}_t}
\mathcal L_{\mathrm{guide}},\\
\widetilde{\Pi}_{U_t}(V_U^{\mathrm{guide}})
&=
\widetilde{\Pi}_{U_t}
\left(
V_U
-
\alpha\nabla_{U_t}
\mathcal L_{\mathrm{guide}}
\right),
\end{aligned}
\end{equation}
where $\alpha\geq0$ controls the guidance strength.

The guided dynamics preserve all spectral constraints enforced by LaSFM: eigenvector orthonormality, the Laplacian nullspace condition, and eigenvalue positivity hold throughout the trajectory for any guidance strength $\alpha$. We formalize it in the following corollary.

\begin{table*}[t]
    \centering
    \caption{$\mathrm{MMD}^2$ ratios for eigenvalue and wavelet statistics,
measuring agreement between generated and test spectral distributions
(lower is better). LaSFM achieves the most consistent performance
across datasets.}
    \label{tab:generation}
    \small
    \setlength{\tabcolsep}{5pt}
    \renewcommand{\arraystretch}{1.15}
    \begin{tabular}{l|cc|cc|cc|cc}
        \hline
        & \multicolumn{2}{c|}{Planar}
        & \multicolumn{2}{c|}{SBM}
        & \multicolumn{2}{c|}{Community-Small}
        & \multicolumn{2}{c}{QM9} \\
        Method
        & Eigenvalue\,\downbetter & Wavelet\,\downbetter
        & Eigenvalue\,\downbetter & Wavelet\,\downbetter
        & Eigenvalue\,\downbetter & Wavelet\,\downbetter
        & Eigenvalue\,\downbetter & Wavelet\,\downbetter \\
        \hline
        SPECTRE & \textbf{3.092} & 26.87 & 14.21 & \underline{1.612} & \underline{1.051} & \underline{2.378} & \underline{2.974} & \underline{15.39}  \\
        GGSD & 25.21 & 397.1 & 3.405 & 82.64 & 2.106 & 30.66 & 19.35 & 45.31\\
        SFMG & 21.35 & \underline{14.95} & \textbf{1.811} & 11.83 & 2.101 & 5.695 & 25.92& 23.16\\
        LaSFM (ours) & \underline{3.839} & \textbf{4.562} & \underline{1.814} & \textbf{2.333} & \textbf{1.000} & \textbf{1.023} & \textbf{1.827} & \textbf{1.561} \\
        \hline
    \end{tabular}
\end{table*}

\begin{Corollary}
\label{cor:guided_preservation}
Suppose the guided dynamics are initialized with
$U_0^\top U_0=I_k$,
$U_0^\top\mathbf{1}=\mathbf{0}$, and
$\widetilde{\Lambda}_0\in\mathbb R^{k\times k}$.
Then, throughout the guided trajectory,
\begin{equation}
U_t^\top U_t=I_k,
\qquad
U_t^\top\mathbf{1}=\mathbf{0},
\qquad
\Lambda_t\succ 0.
\end{equation}
Thus, the guided dynamics preserve all imposed spectral constraints.
\end{Corollary}

\noindent\textit{Proof.}
See Appendix~\ref{app:proofs}.

Corollary~\ref{cor:guided_preservation} shows that the generated eigensystem can be steered toward target properties without sacrificing the spectral constraints enforced by LaSFM.

\section{Related work}

Spectral graph generation models graphs through the eigensystem of
their Laplacian, providing a continuous alternative to directly
generating discrete graph structure. GSDM~\citep{luo2023fast} uses
score-based diffusion to generate eigenvalues while sampling
eigenvectors from the training data, limiting the diversity of the
generated eigensystems. GGSD~\citep{minello2025generating} instead
generates both eigenvalues and eigenvectors using a
DDPM~\citep{ho2020denoising} in Euclidean space, but does not explicitly
preserve the structural constraints of the Laplacian.

Subsequent methods incorporate eigenvector orthonormality.
SPECTRE~\citep{martinkus2022spectre} generates both spectral components
using a GAN~\citep{goodfellow2014generative} and preserves
orthonormality through Stiefel rotation layers, while
SFMG~\citep{huang2025graph} enforces it through flow matching on the
Stiefel manifold. However, neither explicitly preserves two
graph-Laplacian properties: the constant-vector nullspace condition and
positivity of the nonzero eigenvalues. Our work addresses these
remaining constraints while retaining eigenvector orthonormality.



\begin{table*}[h]
    \centering
    \caption{
    $\mathrm{MMD}^2$ ratios for graph statistics (lower is better) and
the joint validity--uniqueness--novelty (VUN) score (higher is better).
LaSFM achieves the most consistent performance across datasets.
    }
    \label{tab:graph_mmd}
    \small
    \setlength{\tabcolsep}{3pt}
    \renewcommand{\arraystretch}{1.15}
    \begin{tabular}{l|cccc|cccc|cccc}
        \hline
        & \multicolumn{4}{c|}{Planar}
        & \multicolumn{4}{c|}{SBM}
        & \multicolumn{4}{c}{Community-Small} \\
        Method
        & Deg.\,\downbetter & Clust.\,\downbetter & Orbit\,\downbetter & V.U.N\,\upbetter
        & Deg.\,\downbetter & Clust.\,\downbetter & Orbit\,\downbetter & V.U.N\,\upbetter
        & Deg.\,\downbetter & Clust.\,\downbetter & Orbit\,\downbetter & V.U.N\,\upbetter \\
        \hline
        SPECTRE
        & \textbf{2.500} & 2.850 & \underline{2.400} & 0.1400
        & 3.000 & 1.670 & \textbf{1.680} & 0.1000
        & 5.910 & 10.87 & 18.97 & \textbf{0.9900} \\

        GGSD
        & 8.500 & \underline{2.800} & 2.800 & 0.1500
        & \underline{2.630} & \textbf{1.660} & 2.650 & 0.1000
        & \underline{1.040} & \underline{2.190} & 10.98 & 0.9500 \\

        SFMG
        & 127.0 & 26.30 & 33.60 & \textbf{0.1800}
        & 32.75 & 2.080 & 3.440 & \textbf{0.2400}
        & 1.790 & 4.860 & \textbf{2.910} & 0.9000 \\

        LaSFM
        & \textbf{2.500} & \textbf{2.730} & \textbf{1.400} & \textbf{0.1800}
        & \textbf{1.250} & \underline{1.780} & \underline{1.880} & \underline{0.2000}
        & \textbf{1.037} & \textbf{1.952} & \underline{9.720} & \underline{0.9700} \\
        \hline
    \end{tabular}
\end{table*}

\section{Experiments}
\label{sec:experiments}

We evaluate LaSFM across the spectral generation pipeline.
First, we assess whether constraint-preserving generation better
matches the data distribution in spectral space (Sec. \ref{sec:exp_eigen}) and after graph
reconstruction (Sec. \ref{sec:exp_graph}). We then evaluate property-guided generation,
testing whether the spectral constraints remain preserved while
steering samples toward target properties (Sec. \ref{sec:exp_guidance}). Finally, we ablate each
constraint to assess its contribution to spectral consistency and
reconstruction (Sec. \ref{sec:exp_ablation}).

\paragraph{Experimental setup.}
We evaluate on the standard graph-generation datasets Planar, SBM,
Community-small~\citep{martinkus2022spectre}, and QM9
\citep{ramakrishnan2014quantum,wu2018moleculenet}.
Following \citet{martinkus2022spectre}, we use 80\%/20\%
training--test splits and compare against
SPECTRE~\citep{martinkus2022spectre},
GGSD~\citep{minello2025generating}, and
SFMG~\citep{huang2025graph}.
We use a common $k$ across methods for spectral evaluation and
the recommended method-specific $k$ for graph generation.
SFMG is a best-effort reproduction because its code is not
publicly available. Further experimental details are provided in
Appendix~\ref{app:experiment-details}, with additional results in
Appendix~\ref{app:additional-results}.


\subsection{Spectral Generation}
\label{sec:exp_eigen}
We first evaluate how well LaSFM captures the spectral
data distribution relative to existing methods.
For all methods, we evaluate eigensystems generated
directly in spectral space, before graph reconstruction.

\paragraph{Metrics.}
We report the ratio of generated--test to train--test
$\mathrm{MMD}^2$ for eigenvalue and graph-wavelet
statistics~\citep{gretton2012kernel,martinkus2022spectre,hammond2011wavelets}.
A lower ratio indicates better agreement with the test distribution, while $1$ represents the empirical train--test discrepancy.
We use the same
number of retained eigenpairs $k$ across methods.


\paragraph{Results.}
LaSFM achieves consistently low eigenvalue and
graph-wavelet $\mathrm{MMD}^2$ ratios across datasets
(Table~\ref{tab:generation}), indicating that it captures
both the Laplacian eigenvalue distribution and the spectral
structure determined jointly by eigenvalues and eigenvectors.
These results suggest that enforcing the Laplacian constraints
provides a useful inductive bias for matching the spectral data
distribution.





\subsection{Graph Generation}
\label{sec:exp_graph}

We next test whether preserving the Laplacian spectral constraints
translates into higher-quality discrete graphs after reconstruction.



\vspace{-3mm}
\paragraph{Graph reconstruction.}
Spectral flow matching generates the eigensystem and thus a low-rank approximation of Laplacian,
$\smash{\widehat L=\widehat U\widehat\Lambda\widehat U^\top}$.
However, it does not directly represent a discrete graph.
Thus, following \citet{huang2025graph}, we train a Euclidean
flow-matching postprocessor that maps $\smash{\widehat L}$
to a continuous adjacency matrix. We then symmetrize and
discretize the output to obtain the graph adjacency matrix $A$.
Reconstruction details are provided in Appendix ~\ref{app:experiment-details}.

\paragraph{Metrics.}
We report $\mathrm{MMD}^2$ ratios for degree, clustering, and
orbit statistics~\citep{minello2025generating}, capturing local
connectivity, neighborhood transitivity, and small-subgraph
structure, respectively. We also report the joint validity--uniqueness--novelty (VUN) score, with validity defined as follows: for Planar, it requires connectedness and planarity; for SBM, we use the SPECTRE criterion~\citep{martinkus2022spectre} to assess consistency with the ground-truth SBM; for Community-small, which lacks a validity criterion, we set validity to 1. Uniqueness measures the fraction of non-isomorphic generated graphs, while novelty measures the fraction not found in the training set.


\paragraph{Results.}
LaSFM achieves the strongest overall graph-level performance,
with consistently low MMD ratios across degree, clustering,
and orbit statistics while maintaining competitive VUN
(Table~\ref{tab:graph_mmd}).
Together with its stronger spectral distribution matching
(Sec.~\ref{sec:exp_eigen}), these results show that LaSFM's
improvements in spectral generation also translate to improved
graph-level distribution matching after reconstruction.

\subsection{Constraint-Preserving Property Guidance}
\label{sec:exp_guidance}

We next investigate whether spectral generation can be steered
toward desired molecular properties while preserving the
spectral constraints and maintaining molecular validity after
reconstruction.

\begin{figure}[b!]
    \centering
    \includegraphics[width=\linewidth]{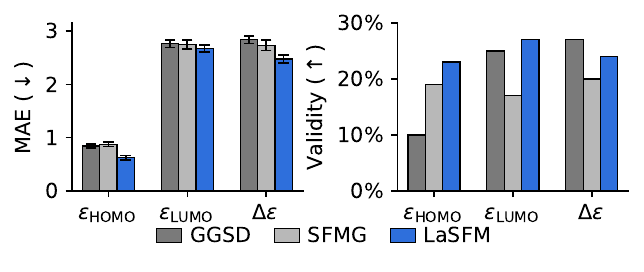}
    \vspace{-5mm}
    \caption{Property-guided QM9 generation evaluated on reconstructed molecules.
    LaSFM achieves the lowest target MAE across all properties while maintaining competitive molecular validity.}
    \label{fig:guidance_reconstructed}
\end{figure}

\paragraph{Setup.}
We evaluate property-guided QM9 generation for three targets:
HOMO ($-5.5$ eV), LUMO ($-1.5$ eV), and the
HOMO--LUMO gap ($5.0$ eV).
For each property, we train a spectral Transformer
$\hat y_\phi$ to predict the property from the eigensystem
and atom types, and guide generation using
$\mathcal{L}_{\mathrm{guide}}
=(\hat y_\phi-y^\star)^2$
as in Eq.~\ref{eq:guided_dynamics}.
We apply the same guidance procedure to LaSFM, GGSD,
and SFMG.

For a controlled comparison, all methods generate 9-node
molecules using all $k=8$ nontrivial Laplacian eigenpairs.
Unlike in Sec.\ref{sec:exp_graph}, we reconstruct
graphs by thresholding the generated Laplacian rather
than using a learned graph predictor. The threshold is selected separately for each method to maximize molecular validity. This isolates the effect
of spectral guidance and avoids a learned reconstruction model
correcting differences between the generated eigensystems. We then use RDKit~\citep{rdkit} to construct molecular structures from the reconstructed graphs.
Guidance strengths ($\alpha$ in Eq.~\ref{eq:guided_dynamics}) are selected to balance target-property accuracy and molecular validity. Further
details are provided in Appendix ~\ref{app:experiment-details}.

\paragraph{Metrics.}
We report the mean absolute error (MAE) between the predicted and target property values (HOMO, LUMO, and HOMO–LUMO gap), along with its standard error. Property values are predicted on valid reconstructed molecules using independently trained directed message-passing neural networks (D-MPNNs)~\citep{yang2019analyzing}. We also report the fraction of valid molecules. A molecule is considered valid if its heavy-atom graph is connected and RDKit successfully performs bond-order assignment and sanitization. Finally, we measure spectral constraint violations through orthonormality error $\mathcal{E}_{\mathrm{orth}}$, nullspace error $\mathcal{E}_{\mathrm{null}}$, and eigenvalue positivity error $\mathcal{E}_{\lambda}$. All metrics are evaluated over 1,000 generated samples, with further definitions in Appendix~\ref{app:experiment-details}.

\paragraph{Results.}
As shown in Fig.~\ref{fig:guidance_reconstructed}, LaSFM achieves
the lowest target MAE on reconstructed molecules with highest
validity. Consistent with Corollary~\ref{cor:guided_preservation},
it also has zero $\mathcal{E}_{\mathrm{orth}}$,
$\mathcal{E}_{\mathrm{null}}$, and $\mathcal{E}_{\lambda}$ for LUMO
(Table~\ref{tab:guided_constraint_errors}), while the baselines
violate these constraints. This suggests that constraint-preserving
guidance improves target alignment without degrading the generated
eigensystem. Results for other properties are in
Appendix~\ref{app:additional-results}.

\begin{figure}
    \centering
    \includegraphics[width=0.9\linewidth]{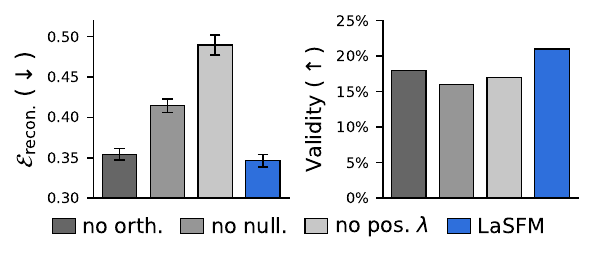}
    \vspace{-5mm}
    \caption{Ablation of the Laplacian constraints on QM9 generation. LaSFM achieves the lowest reconstruction error and highest validity, showing the importance of all three constraints.}
    \label{fig:ablation}
\end{figure}

\subsection{Ablation}
\label{sec:exp_ablation}

Finally, we ablate each Laplacian constraint and evaluate reconstruction consistency and molecular validity.

\paragraph{Setup.}
We use the same QM9 setting as
Sec.~\ref{sec:exp_guidance}. We ablate i) the nullspace
constraint by removing the projection onto the orthogonal
complement of $\mathbf{1}$, ii) orthonormality by replacing
Stiefel flow matching with Euclidean flow matching, and
iii) eigenvalue positivity by removing the positive-eigenvalue
mapping in Eq.~\ref{eq:lambda_map}. All other settings
are kept fixed.

\paragraph{Metrics.}
We measure reconstruction consistency using the Laplacian reconstruction
error $\mathcal{E}_{\mathrm{recon}}$, defined as the discrepancy
between the generated spectral Laplacian and the Laplacian of the
thresholded reconstructed graph. We also report molecular validity,
requiring a connected heavy-atom graph and successful RDKit bond-order
assignment and sanitization. Results are reported over 1,000 samples.
Lower $\mathcal{E}_{\mathrm{recon}}$ indicates better agreement between the
generated eigensystem and reconstructed graph, while higher validity
indicates better molecular reconstruction. Exact definitions are provided in
Appendix~\ref{app:experiment-details}.
\begin{table}[!h]
    \centering
    \caption{
    Spectral constraint errors under LUMO guidance.
    LaSFM preserves all three constraints.
    }
    \label{tab:guided_constraint_errors}
    \small
    \setlength{\tabcolsep}{6pt}
    \renewcommand{\arraystretch}{1.05}

    \vspace{1mm}
    \begin{tabular}{lccc}
        \hline
        Method
        & $\mathcal E_{\mathrm{orth}}\downarrow$
        & $\mathcal E_{\mathrm{null}}\downarrow$
        & $\mathcal E_{\lambda}\downarrow$ \\
        \hline
        GGSD
        & 0.1062
        & 0.0124
        & 0.0003 \\

        SFMG
        & 0.0000
        & 0.2493
        & 0.0146 \\

        LaSFM
        & \textbf{0.0000}
        & \textbf{0.0000}
        & \textbf{0.0000} \\
        \hline
    \end{tabular}
\end{table}

\paragraph{Results.}
The full LaSFM model outperforms every ablated variant, achieving the
lowest reconstruction error $\mathcal{E}_{\mathrm{recon}}$ and highest
molecular validity, as shown in Fig.~\ref{fig:ablation}. Removing any Laplacian constraint increases
the correction required during reconstruction, reducing consistency
between the generated spectral representation and the reconstructed
graph.

\begin{figure}
    \centering
    \includegraphics[width=\linewidth]{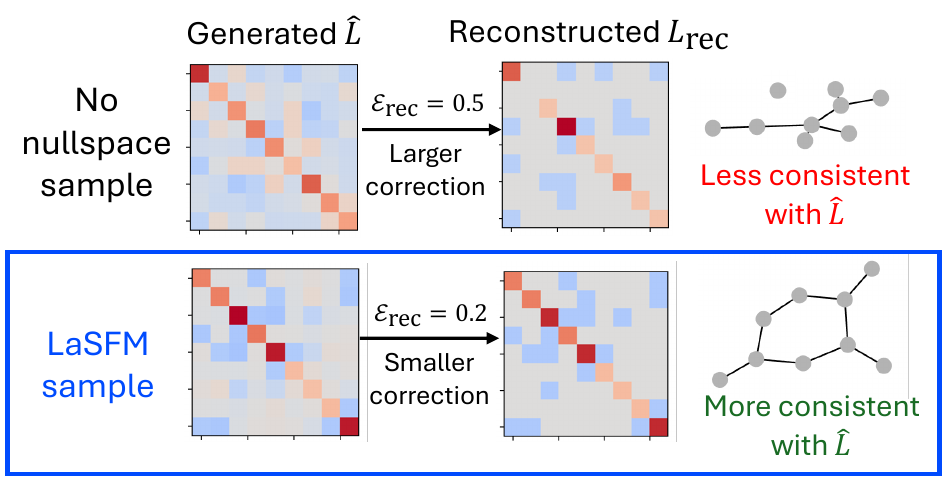}
    \vspace{-5mm}
    \caption{
Compared with variant without nullspace constraint, LaSFM generates Laplacians that require smaller corrections, thereby producing graphs that are more consistent with the generated Laplacians.
}
    \label{fig:heatmap}
\end{figure}

Figure~\ref{fig:heatmap} illustrates the effect of the nullspace constraint. Compared with LaSFM, the variant without this constraint produces a larger discrepancy between the generated spectral Laplacian and the Laplacian of the reconstructed graph, as captured by $\mathcal{E}_{\mathrm{recon}}$. This helps explain why preserving the constraints improves LaSFM's reconstruction performance.

\section{Conclusion}

Our work focuses on two graph-Laplacian properties not addressed by
existing spectral generative methods: the constant-vector nullspace
condition and positivity of the nonzero eigenvalues. We characterize
the corresponding constrained Stiefel geometry and propose LaSFM,
which preserves these properties together with eigenvector
orthonormality during generation and property guidance. We empirically demonstrate improved spectral matching, graph reconstruction, and target alignment, while ablations show that removing any constraint degrades reconstruction.

There are, however, two remaining limitations. First, the enforced spectral constraints do not fully characterize a combinatorial Laplacian. In particular, truncated eigensystems do not guarantee the required nonpositive off-diagonal entries, which LaSFM does not enforce. Second, converting the generated eigensystem into a discrete graph still requires a reconstruction procedure, which is beyond the scope of our work.

\subsection*{Acknowledgements}

KK was supported by the Alexander von Humboldt Foundation
(Humboldt Research Award), JSPS KAKENHI
(JP26K21777 and JP25K24465), and JST ASPIRE
(JPMJAP2302). YJ was supported by the National IT Industry Promotion Agency (NIPA) grant funded by the Korean government (MSIT) (Physical AI SDF-Specialized Foundation Model for Autonomous Factory Operations, RS-2026-25621588) and the National Research Foundation of Korea (NRF) grant funded by the Korean government (MSIT) (RS-2026-25588527).

\bibliography{references}

\clearpage
\appendix
\thispagestyle{empty}

\onecolumn
\aistatstitle{Supplements}

\appendix

\section{Background}\label{app:background}

\subsection{Flow Matching in Spectral Space}
\label{app:spectral-flow}

\paragraph{Spectral graph generation via flow matching.}

Flow matching\citep{lipman2022flow,albergo2022building,liu2023flow} is a generative modeling method that learns a time-dependent vector field that transports samples from a simple base distribution at $t=0$ to the data distribution at $t=1$. Spectral graph generation can leverage
flow matching to generate the eigenvectors $U$ and eigenvalues $\Lambda$, respectively~\citep{huang2025graph}.

For the time index $t\in[0,1]$, let $U_t$ and $\Lambda_t$ denote the eigenvector and
eigenvalue matrices at time $t$, respectively. Then, their evolutions are characterized 
by the following vector fields
\begin{equation}    
    \dot U_t:= \frac{\mathrm{d}U_t}{\mathrm{d}t},
    \qquad  \qquad  
    \dot \Lambda_t :=  
    \frac{\mathrm{d}\Lambda_t}{\mathrm{d}t},
\end{equation}
and $U_t$ and $\Lambda_t$ are specified as follows:
\begin{equation}
\begin{aligned}
    U_t
    = U_0+\int_0^t \dot U_s\,\mathrm{d}s, \qquad
    \Lambda_t
    = \Lambda_0+\int_0^t
    \dot \Lambda_s  \,\mathrm{d}s,
\end{aligned}
\end{equation}
where $U_0$ and $\Lambda_0$ are sampled from the base distributions.
The $U_1$ and $\Lambda_1$ follow the corresponding data distributions.

Since the true vector fields $\dot{U}_t$ and $\dot{\Lambda}_t$ are unknown, Flow Matching approximates them using neural networks.
\begin{equation}
V_U(t,U_t), \qquad
V_\Lambda(t,\Lambda_t),
\end{equation}
Given the current time and state, $V_U$ and $V_\Lambda$ predict the corresponding velocities, thereby approximating $\dot{U}_t$ and $\dot{\Lambda}_t$, respectively.

\subsection{Stiefel Manifold Geometry} \label{app:stiefel-background}

Here, we provide background on the Stiefel manifold and its exponential and logarithmic maps necessary to understand our method. Our description follows \cite{cheng2025stiefel}; further details are provided in \cite{edelman1998geometry,zimmermann2022computing}.

\paragraph{Stiefel manifold}

The Stiefel manifold $\mathrm{St}(n,k)$ denotes set of $n \times k$ orthonormal matrices defined as
\begin{equation}
\mathrm{St}(n,k) = \{U \in \mathbf{R}^{n \times k} \mid U^\top U = I_k\}.
\end{equation}
It is a smooth manifold of dimension $nk-k(k+1)/2$. Intuitively, each element of $\mathrm{St}(n,k)$ represents an ordered collection of $k$ orthonormal vectors in $\mathbb{R}^n$.

\paragraph{Tangent space}

The tangent space of $\mathrm{St}(n,k)$ at a point $U$ is defined as
\begin{equation}
\label{eq:stiefel_tanspace}
T_U\mathrm{St}(n,k)
=
\left\{
\Delta\in\mathbb{R}^{n\times k}
\mid
U^\top\Delta+\Delta^\top U=\vzero
\right\}.
\end{equation}
This set is obtained by differentiating the orthonormality constraint defining
$\mathrm{St}(n,k)$.

\paragraph{Exponential map}

Given an initial point $U_0\in\mathrm{St}(n,k)$ and a tangent vector
$\Delta\in T_{U_0}\mathrm{St}(n,k)$, we can define geodesic $\gamma(t)$ for
$t\in[0,1]$ with the following condition
\begin{equation}
\gamma(0)=U_0, \qquad \dot{\gamma}(0)=\Delta.
\end{equation}
The exponential map sends the point $t\Delta \in T_{U_0}\mathrm{St}(n,k)$ to the point
\begin{equation}
\gamma(t)=\operatorname{Exp}_{U_0}(t\Delta)\in\mathrm{St}(n,k).
\end{equation}
We denote the geodesic point $\gamma(t)$ by $U_t$.

\paragraph{Logarithmic map}
The logarithmic map identifies the tangent vector $\Delta$ of a geodesic connecting two points on the Stiefel manifold. Given and initial point $U_0\in\mathrm{St}(n,k)$ and an endpoint $U_1\in\mathrm{St}(n,k)$, let $\gamma(t)$ be a geodesic satisfying
\[
U_0=\gamma(0), \qquad U_1=\gamma(1).
\]
Its initial velocity $\dot{\gamma}(0)\in T_{U_0}\mathrm{St}(n,k)$ can be obtained from the logarithmic map:
\begin{equation}
\dot{\gamma}(0)
=
\operatorname{Log}_{U_0}(U_1).
\end{equation}
Thus, for a fixed base point $U_0$, $\operatorname{Log}_{U_0}$ maps the endpoint $U_1$ to the tangent vector $\dot{\gamma}(0)$ in $T_{U_0}\mathrm{St}(n,k)$. We denote the tangent vector $\dot{\gamma}(0)$ as $\Delta$.

\begin{figure}[t]
    \centering
    \includegraphics[width=0.3\textwidth]{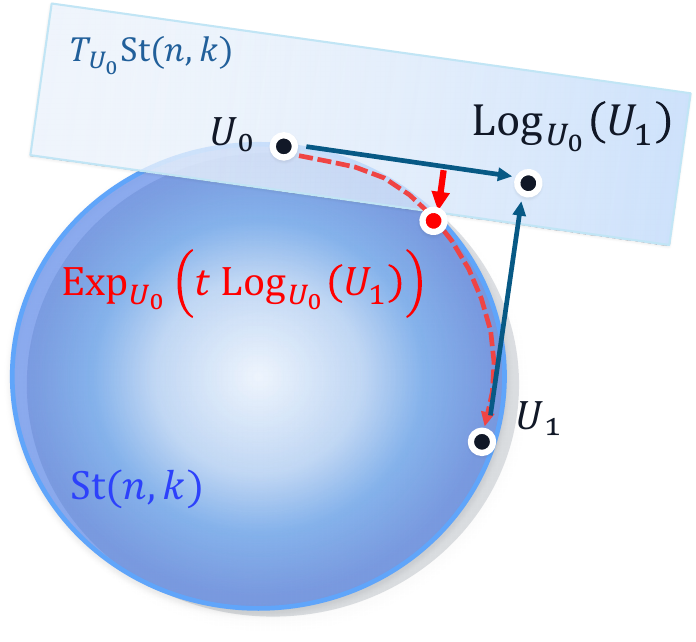}
    \caption{Illustration of exponential and logarithmic maps.}
    \label{fig:exp-log-map_main}
\end{figure}

\paragraph{Geodesic representation}

Therefore, the geodesic can be expressed using the exponential and logarithmic
maps as
\begin{equation}
U_t
=
\operatorname{Exp}_{U_0}
\left(
t\operatorname{Log}_{U_0}(U_1)
\right),
\qquad t\in[0,1],
\end{equation}
where $U_0=\gamma(0)$ and $U_1=\gamma(1)$.

\begin{figure*}[t]
\centering

\begin{minipage}[t]{0.485\textwidth}
\vspace{0pt}

\hrule height 0.8pt
\vspace{1mm}

\refstepcounter{algorithm}
\label{alg:stiefelexp}
\noindent\textbf{Algorithm \thealgorithm: Exponential Map $\gamma(t)$}

\vspace{1mm}
\hrule height 0.4pt
\vspace{2mm}

\begin{algorithmic}[1]
\Require
\parbox[t]{0.82\linewidth}{
Base $U_0\in\mathrm{St}(n,k)$, time $t\in(0,1)$,
and initial direction $\Delta\in T_{U_0}\mathrm{St}(n,k)$
}

\vspace{2mm}

\State Compute the QR decomposition
\[
QR
\leftarrow
(I_n-U_0U^\top_0)\Delta,
\]
where $Q\in\mathbb{R}^{n\times k}$ and
$R\in\mathbb{R}^{k\times k}$

\State Construct the block matrix
\[
A
\leftarrow
\begin{pmatrix}
U^\top_0\Delta & -R^\top\\
R & \mzeros_{k\times k}
\end{pmatrix}
\]

\State Compute the matrix exponential
\[
\begin{pmatrix}
M(t) & \cdots\\
N(t) & \cdots
\end{pmatrix}
\leftarrow
\exp(tA)
\]

\State Compute the geodesic point
\[
\gamma(t)
\leftarrow
U_0M(t)+QN(t)
\]

\State \Return $\gamma(t)\in\mathrm{St}(n,k)$
\end{algorithmic}

\vspace{1mm}
\hrule height 0.8pt

\end{minipage}
\hfill
\begin{minipage}[t]{0.485\textwidth}
\vspace{0pt}

\hrule height 0.8pt
\vspace{1mm}

\refstepcounter{algorithm}
\label{alg:stiefellog}
\noindent\textbf{Algorithm \thealgorithm: Logarithm Map
$\operatorname{Log}_{U_0}(U_1)$}

\vspace{1mm}
\hrule height 0.4pt
\vspace{2mm}

\begin{algorithmic}[1]
\Require
\parbox[t]{0.82\linewidth}{
Base $U_0\in\mathrm{St}(n,k)$ and target
$U_1\in\mathrm{St}(n,k)$
}

\vspace{2mm}

\State $M\leftarrow U_0^\top U_1$
\State $QN\leftarrow U_1-U_0M$

\State Choose $X,Y\in\mathbb{R}^{k\times k}$ such that
\[
V
\leftarrow
\begin{pmatrix}
M & X\\
N & Y
\end{pmatrix},
\qquad
V^\top V=I_{2k}
\]

\For{$j=1,\ldots,20$}
    \State
    \[
    \begin{pmatrix}
    A & -B^\top\\
    B & C
    \end{pmatrix}
    \leftarrow
    \log(V)
    \]

    \State
    $
    S\leftarrow
    \frac{1}{12}BB^\top-\frac{1}{2}I_k
    $

    \State Solve
    $
    C=S\Gamma+\Gamma S
    $

    \State
    $
    \Phi\leftarrow\exp(\Gamma)
    $

    \State
    \[
    V
    \leftarrow
    V
    \begin{pmatrix}
    I_k & \mzeros_{k\times k}\\
    \mzeros_{k\times k} & \Phi
    \end{pmatrix}
    \]
\EndFor

\State \Return
$\operatorname{Log}_{U_0}(U_1)=U_0A+QB$

\end{algorithmic}

\vspace{1mm}
\hrule height 0.8pt

\end{minipage}

\end{figure*}

\paragraph{Computing Exponential and Logarithmic Maps}
Although the exponential and logarithmic maps are mathematically well defined, their computation is nontrivial. Fortunately, \citep{edelman1998geometry,zimmermann2022computing} provide practical algorithms for computing the exponential and logarithmic maps, respectively.

\vspace{1mm}
\textbf{Algorithm~\ref{alg:stiefelexp}} describes the computation of the exponential map, which takes an initial point $U_0$ and a tangent vector $\Delta\in T_{U_0}\mathrm{St}(n,k)$ and returns the corresponding geodesic point $\operatorname{Exp}_{U_0}(t\Delta)$ on the Stiefel manifold.

\vspace{1mm}
\textbf{Algorithm~\ref{alg:stiefellog}} computes the logarithmic map, which takes an initial point $U_0$ and an endpoint $U_1$ on the Stiefel manifold and returns the corresponding tangent vector
$\operatorname{Log}_{U_0}(U_1)\in T_{U_0}\mathrm{St}(n,k)$.
We adopt the implementation of \cite{cheng2025stiefel}, which modifies the Stiefel logarithm algorithm of \cite{zimmermann2022computing} for computational efficiency. Its dominant cost is $O(nk^2)$, arising from the QR decomposition and initial matrix multiplications.

\section{Theoretical Details}
\label{app:theory}

\subsection{Nullspace-Constrained Stiefel Geometry}
\label{app:nullspace-geometry}

Recall the nullspace-constrained Stiefel manifold
\[
\mathcal M_{n,k}
=
\left\{
U\in\operatorname{St}(n,k):
U^\top\mathbf{1}=\mathbf{0}
\right\}.
\]

Its tangent space is
\[
T_U\mathcal M_{n,k}
=
\left\{
\Delta\in\mathbb R^{n\times k}:
U^\top\Delta+\Delta^\top U=0,\;
\Delta^\top\mathbf{1}=\mathbf{0}
\right\}.
\]
Thus, tangent velocities preserve both eigenvector
orthonormality and orthogonality to the constant vector.

The manifold $\mathcal M_{n,k}$ is isometric to
$\operatorname{St}(n-1,k)$ and is a totally geodesic
submanifold of $\operatorname{St}(n,k)$. Consequently,
geodesics initialized in $\mathcal M_{n,k}$ with tangent
velocity in $T_U\mathcal M_{n,k}$ remain in
$\mathcal M_{n,k}$.

We prove these properties below.

\subsection{Proof of Propositions}\label{app:proofs}


We first present Lemmas~1--3, which describe a
reduced-dimensional representation of the Stiefel geodesic path induced from the nullspace constraint. We then use this representation to
prove Propositions~1--4.

\begin{lemma} \label{lem:nullspace_reduction}
If $U \in \operatorname{St}(n,k)$ and 
$V \in T_{{U}}\operatorname{St}(n,k)$
satisfy the nullspace constraints 
\[ 
 U  ^\top \allone= \mzeros, 
\qquad
 V  ^\top \allone = \mzeros, 
\] 
then there exists a basis matrix $B \in \mathbb R^{n\times (n-1)}$ whose columns form an orthonormal basis for $\nallone_\perp$, and reduced representations,
\[ 
\widetilde{U} \in \operatorname{St}(n-1,k), 
\qquad 
\widetilde{V} \in T_{\widetilde{U}}\operatorname{St}(n-1,k),
\]
where $U=B\widetilde U$ and $V=B\widetilde V$

\begin{proof}

%

Since the dimension of \(\nallone_\perp\) is \(n-1\), the orthonormal basis matrix \(B\in\mathbb R^{n\times(n-1)}\) for
\(\nallone_\perp\) exists and satisfies
\[
B^\top B=I_{n-1},
\qquad
\nallone^\top B=\vzero.
\]
 We will now show that there exists $\widetilde{U} \in \operatorname{St}(n-1,k)$. Let us consider any
\[
U\in\operatorname{St}(n,k)\cap
\{U:\nallone^\top U=\vzero\},
\]
meaning that all columns of \(U\) belong to \(\nallone_\perp\). Then, there exists
a corresponding
\[
\widetilde U\in\mathbb R^{(n-1)\times k} \quad s.t \quad
U=B\widetilde U
\]
because \(B\) is an orthonormal basis matrix for \(\nallone_\perp\).

Using the \(U\in\operatorname{St}(n,k)\), the following holds
\[
I_k
=
U^\top U
=
\widetilde U^\top B^\top B\widetilde U
=
\widetilde U^\top\widetilde U,
\]
which implies that
\(\widetilde U\in\operatorname{St}(n-1,k)\).

\vspace{3mm}
Next, we will show there exists $\widetilde{V} \in T_{\widetilde U}\operatorname{St}(n-1,k)$.
Similarly, let us consider any 
\[
V\in T_U\operatorname{St}(n,k) \cap  \{V:\nallone^\top V=\vzero\}.
\] 
Then, there exists a corresponding
\[
\widetilde V\in\mathbb R^{(n-1)\times k}
\quad \text{s.t.} \quad
V=B\widetilde V,
\]
because \(B\) is an orthonormal basis matrix for \(\nallone_\perp\).

Moreover, as
\[
U^\top V+V^\top U=0
\]
holds due to \(V\in T_U\operatorname{St}(n,k)\), substituting
\(U=B\widetilde U\) and \(V=B\widetilde V\) into the equation gives
\[
\widetilde U^\top\widetilde V
+
\widetilde V^\top\widetilde U
=
0,
\]
which implies
\[
\widetilde V
\in
T_{\widetilde U}\operatorname{St}(n-1,k).
\]

\end{proof}
\end{lemma}

\begin{lemma}[Corollary~2.2 in \citet{edelman1998geometry}]
For $U \in \operatorname{St}(n,k)$ and
$V \in T_U\operatorname{St}(n,k)$,
the geodesic update on the Stiefel manifold from $U$ in the direction $V$, with step size $\delta>0$, is expressed by exponential map
\[
\operatorname{Exp}_{U}
\!\left(\delta V\right)
=
\begin{bmatrix}
U & Q
\end{bmatrix}
\exp\left(
\delta
\begin{bmatrix}
A & -R^\top\\
R & 0
\end{bmatrix}
\right)
\begin{bmatrix}
I_k\\
0
\end{bmatrix},
\]
where $
A = U^\top V$. The \(Q\) and \(R\) are given by the QR decomposition
\[
QR = (I-UU^\top)V.
\]

\begin{proof}
See Corollary~2.2 in \citet{edelman1998geometry}.
\end{proof}
\end{lemma}

\begin{lemma}
    
If $U \in \operatorname{St}(n,k)$ and 
$V \in T_{U}\operatorname{St}(n,k)$
satisfy the nullspace constraints 
\[
U ^\top \allone = \mathbf{0}, 
\qquad
V ^\top \allone = \mathbf{0}, 
\] 
then the Stiefel exponential map
$\operatorname{Exp}_{U}\!\left(\delta V\right)$
can also be expressed as
\[
\operatorname{Exp}_{U}\!\left(\delta V\right)
=
B\,
\operatorname{Exp}_{\widetilde U}
\!\left(\delta \widetilde V\right),
\]
where $B\in\mathbb R^{n\times(n-1)}$ denotes the matrix whose columns form an orthonormal basis for $\nallone_\perp$. The $\widetilde U \in \operatorname{St}(n-1,k) $ and $\widetilde V \in T_{\widetilde U}\operatorname{St}(n-1,k) $ denote the orthonormal matrix satisfying $U=B\widetilde U$ and $V=B\widetilde V$.

\begin{proof}
Consider that 
for $U \in \operatorname{St}(n,k)$ and $V \in T_{U}\operatorname{St}(n,k)$ satisfying the condition
$\nallone^\top U = \mathbf{0}$
and
$
\nallone^\top V = \mathbf{0}
$, there exist
$\widetilde U \in \operatorname{St}(n-1,k)$ and $\widetilde V \in T_{\widetilde U }\operatorname{St}(n-1,k)$ satisfying
\[
U=B\widetilde U,
\qquad
V=B\widetilde V,
\]
due to Lemma 1. Also, due to Lemma 2, the exponential map is given as
\[
\operatorname{Exp}_{U}
\!\left(\delta V\right)
=
\begin{bmatrix}
U & Q
\end{bmatrix}
\exp\left(
\delta 
\begin{bmatrix}
A&-R^\top\\
R&0
\end{bmatrix}
\right)
\begin{bmatrix}
I_k\\
0
\end{bmatrix}, 
\]
where $
A= U^\top V
$ and $Q$ is given from QR decomposition of $QR = (I-UU^{\top}) V$.

\vspace{1mm}
Then, substituting $U=B\widetilde U$ and $V=B\widetilde V$ into 
$A$ in the exponential map  gives
\[
\begin{aligned}
A
=
U^\top V
=
\widetilde U^\top B^\top B\widetilde V
=
\widetilde U^\top\widetilde V
=
\widetilde A.
\end{aligned}
\]
Also, when $\widetilde Q$ and $\widetilde  R$ are given from the QR decomposition
$ \widetilde Q \widetilde  R =
\left(I-\widetilde U\widetilde U^\top\right)\widetilde V$, the following holds
\[
\begin{aligned}
(I-UU^\top)V &=
\left(I-B\widetilde U\widetilde U^\top B^\top\right)
B\widetilde V \\
&=
B\left(I-\widetilde U\widetilde U^\top\right)\widetilde V
= B \widetilde Q  \widetilde  R.
\end{aligned}
\]
Thus, for the $QR$ decomposition of $(I-UU^\top)V$, we can set 
\[
\begin{aligned}
Q= B\widetilde Q, \qquad R = \widetilde R
\end{aligned}
\]
such that $QR = (I-UU^\top)V $.This is because $B\widetilde Q$ is orthonormal matrix, as
\[
\begin{aligned}
(B\widetilde Q)^\top(B\widetilde Q) = 
\widetilde Q^\top\widetilde Q 
=
I_k,
\end{aligned}
\]
and $\widetilde R$ is an upper triangular matrix. 

Therefore, the following holds
\[
\begin{aligned}
\operatorname{Exp}_{U}\!\left(\delta V\right)
&=
\begin{bmatrix}
U &  Q
\end{bmatrix}
\exp\left(
\delta
\begin{bmatrix}
A & -R^\top\\
R & 0
\end{bmatrix}
\right)
\begin{bmatrix}
I_k\\
0
\end{bmatrix} \\
&=
B
\begin{bmatrix}
\widetilde U & \widetilde Q
\end{bmatrix}
\exp\left(
\delta
\begin{bmatrix}
\widetilde A & -\widetilde R^\top\\
\widetilde R & 0
\end{bmatrix}
\right)
\begin{bmatrix}
I_k\\
0
\end{bmatrix} \\
&=
B\,
\operatorname{Exp}_{\widetilde U}
\!\left(\delta\widetilde V\right).
\end{aligned}
\]

\end{proof}

\end{lemma}


\setcounter{Proposition}{0}

\begin{Proposition}
    For a Laplacian matrix $L$, let
    $U\in\operatorname{St}(n,k)$ and let $\Lambda$ be a diagonal matrix
    with positive eigenvalues such that $
    L=U\Lambda U^\top$.    
    Then,
    \vspace{1mm}
    \[
    L\allone =\mzeros
    \quad\text{if and only if}\quad
    U^{\top} \allone=\mzeros.
    \]

    \begin{proof}
    The implication $(\Leftarrow)$ is trivial. Thus, we only need to prove  that
    \[
    \allone^\top L=\mzeros
    \quad\Longrightarrow\quad
    \allone^\top U=\mzeros.
    \]
    Suppose that $\allone^\top L=\mzeros$ due to $L^{\top} = L$. Using the eigendecomposition $L = U\Lambda U^\top$, and right-multiplying both sides by $U$ gives
    \[
    \begin{aligned}
    \mzeros
    &=
    \allone^\top L U 
    =
    \allone^\top U\Lambda,
    \end{aligned}
    \]
    where we use $U^\top U=I_k$.
    Since $\Lambda$ has strictly positive diagonal entries, it is invertible.
    Therefore,
    \[
    \allone^\top U=\mzeros.
    \]
    \end{proof}
\end{Proposition}

\begin{Proposition}
Let $U_0,U_1\in\operatorname{St}(n,k)$ denote Laplacian eigenvector matrices sampled from the base distribution and the data distribution, respectively. Suppose
\[
U_0^\top\allone=\vzero,
\qquad
U_1^\top\allone=\vzero,
\]
belong to the same connected component and that the minimizing Stiefel geodesic connecting
$U_0$ and $U_1$ is unique. Then, for $t\in[0,1]$, the
geodesic path $U_t$ satisfies
\[
U_t^\top\allone=\vzero.
\]
\begin{proof}
By Lemma~1, write
\[
U_0=B\widetilde U_0,
\qquad
U_1=B\widetilde U_1,
\]
where
$\widetilde U_0,\widetilde U_1
\in\operatorname{St}(n-1,k)$.

Define
\[
\widetilde V_0
=
\operatorname{Log}_{\widetilde U_0}(\widetilde U_1),
\qquad
V_0=B\widetilde V_0.
\]
Then $V_0\in T_{U_0}\operatorname{St}(n,k)$ and
$V_0^\top\allone=\vzero$.

By Lemma~3,
\[
\operatorname{Exp}_{U_0}(tV_0)
=
B\operatorname{Exp}_{\widetilde U_0}
(t\widetilde V_0)
\]
defines a minimizing Stiefel geodesic connecting
$U_0$ and $U_1$. By uniqueness, this is the geodesic
$U_t$ in the proposition. Therefore,
\[
U_t
=
B\operatorname{Exp}_{\widetilde U_0}
(t\widetilde V_0).
\]
Since $\nallone^\top B=0$,
\[
\nallone^\top U_t=\mzeros.
\]

\end{proof}
\end{Proposition}

\paragraph{Uniqueness assumption.}
Proposition~\ref{prop:nullspace_geodesic} assumes that the
minimizing Stiefel geodesic between $U_0$ and $U_1$ is unique.
Under the continuous sampling of $U_0$, endpoint pairs admitting
multiple minimizing geodesics form a measure-zero set.
We therefore assume uniqueness throughout and exclude these
degenerate cases.

\paragraph{Connected components.}
The nullspace-preserving geodesic construction requires
$U_0$ and $U_1$ to lie in the same connected component of
$\mathcal{M}_{n,k}$. For $k<n-1$, $\mathcal{M}_{n,k}$ is
connected, so this condition is automatic. When $k=n-1$,
however, $\mathcal{M}_{n,k}$ has two connected components,
corresponding to the two possible orientations of an
orthonormal basis of $\mathbf{1}^{\perp}$.

Since $U_0$ and $U_1$ are both orthonormal bases of
$\mathbf{1}^{\perp}$ in this case, the sign of
$\det(U_0^\top U_1)$ determines whether they have the same
orientation. If
\[
\det(U_0^\top U_1)=-1,
\]
we flip the sign of one column of $U_1$ by setting
\[
U_1 \leftarrow U_1D,
\qquad
D=\operatorname{diag}(-1,1,\ldots,1).
\]
This moves $U_1$ to the same connected component as $U_0$
without changing the represented Laplacian. Indeed, since
$D$ is diagonal with entries $\pm1$,
\[
D\Lambda D=\Lambda,
\]
and hence
\[
(U_1D)\Lambda(U_1D)^\top
=
U_1\Lambda U_1^\top.
\]

\begin{Proposition}
Suppose $U_0 \in\operatorname{St}(n,k)$ and $\dot U_0 \in T_{U_0}\operatorname{St}(n,k)$, then
there exists a geodesic path $U_t \in\operatorname{St}(n,k)$ for $t\in[0,1]$.  Moreover, if $U_0$ and the velocity $\dot U_t$ satisfies
\[
U_0^{\top} \allone= \mzeros, \qquad \dot U_t ^{\top} \allone=\mzeros,
\]
then $U_t$ satisfies
\[
 U_t ^{\top} \allone =\mzeros.
\]
\begin{proof}
For any point $U_0\in\operatorname{St}(n,k)$ and tangent vector
\[
\dot U_0\in T_{U_0}\operatorname{St}(n,k),
\]
there exists a unique geodesic $\gamma(t)$ satisfying
\[
\gamma(0)=U_0,
\qquad
\dot{\gamma}(0)=\dot U_0.
\]
By the definition of the Riemannian exponential map, this geodesic is expressed as
\[
\gamma(t)
=
\operatorname{Exp}_{U_0}\!\left(t\dot U_0\right).
\]
Further details can be found in Section~5.4 of
\cite{absil2008optimization}. Here, we denote $U_t:=\gamma(t)$.

Then, by the fundamental theorem of calculus, 
\[ 
U_t = U_0 + \int_0^t \dot U_s\,ds 
\]
is given and 
multiplying both sides by $\allone^\top$ gives
\[ 
\begin{aligned}
\allone^\top U_t &= \allone^\top U_0 + \int_0^t \allone^\top\dot U_s\,ds = \mzeros. 
\end{aligned} 
\] due to  $\allone^\top U_0=\mathbf{0}$ and $\allone^\top\dot U_s = \mzeros$ for $s\in[0,1]$.

\end{proof}
\end{Proposition}

\begin{Proposition}
\label{app:proof-projected-vector-field}
For $U_t \in \operatorname{St}(n,k)$ satisfying $U_t^\top \mathbf{1} =0$, the projected velocity
$\widetilde{\Pi}(V_U)$ defined in Eq.~\eqref{eq:pproj} satisfies
\[
\widetilde{\Pi}(V_U)^\top\allone=\mzeros,
\qquad
U_t^\top\widetilde{\Pi}(V_U)
+
\widetilde{\Pi}(V_U)^\top U_t
=0.
\]
Therefore, $\widetilde{\Pi}(V_U)$ preserves both the
nullspace constraint and the Stiefel tangent-space condition.
\begin{proof}
The projection onto $\mathbf{1}^\perp$ is given in Eq. \ref{eq:project_centered} as,
\[
P=I_n-\frac{1}{n}\mathbf{1}\mathbf{1}^\top,
\]
and define
\[
W=PV_U,
\qquad
S=\operatorname{sym}(U_t^\top W).
\]
Then the projected velocity in Eq.~\eqref{eq:pproj} is
\[
\widetilde{\Pi}(V_U)
=
W-U_tS.
\]

Since $P$ projects onto the orthogonal complement of
$\mathbf{1}$,
\[
W^\top\mathbf{1}=\mathbf{0}.
\]
Together with the assumption
$U_t^\top\mathbf{1}=\mathbf{0}$, this gives
\[
\begin{aligned}
\widetilde{\Pi}(V_U)^\top\mathbf{1}
&=
(W-U_tS)^\top\mathbf{1}\\
&=
\mathbf{0}.
\end{aligned}
\]

Moreover, since $U_t\in\operatorname{St}(n,k)$,
$U_t^\top U_t=I_k$. Therefore,
\[
\begin{aligned}
&U_t^\top\widetilde{\Pi}(V_U)
+
\widetilde{\Pi}(V_U)^\top U_t\\
&=
U_t^\top W+W^\top U_t
-
S-S^\top\\
&=
U_t^\top W+W^\top U_t
-
2\operatorname{sym}(U_t^\top W)\\
&=
0.
\end{aligned}
\]

Hence, $\widetilde{\Pi}(V_U)$ satisfies both the Stiefel
tangent-space condition and
$\widetilde{\Pi}(V_U)^\top\mathbf{1}=\mathbf{0}$.
\end{proof}
\end{Proposition}

\setcounter{Corollary}{0}
\begin{Corollary}
Suppose the guided dynamics are initialized with
$U_0^\top U_0=I_k$,
$U_0^\top\mathbf{1}=\mathbf{0}$, and
$\widetilde{\Lambda}_0\in\mathbb R^{k\times k}$.
Then, throughout the guided trajectory,
\begin{equation}
U_t^\top U_t=I_k,
\qquad
U_t^\top\mathbf{1}=\mathbf{0},
\qquad
\Lambda_t\succ 0,
\end{equation}
so the guided dynamics preserve all imposed spectral constraints.
\begin{proof}
By Proposition~\ref{prop:nullspace_projection}, the guided
eigenvector velocity
$\widetilde{\Pi}_{U_t}(V_U^{\mathrm{guide}})$
satisfies both the Stiefel tangent-space and nullspace velocity
conditions. Therefore, $U_t^\top U_t=I_k$ is preserved along the
guided trajectory, while Proposition~\ref{prop:nullspace_velocity}
implies preservation of $U_t^\top\mathbf{1}=\mathbf{0}$.

For the eigenvalues, the mapping in
Eq.~\eqref{eq:lambda_map} yields strictly positive, ordered
eigenvalues for every finite $\widetilde{\Lambda}_t$.
Therefore, all three constraints are satisfied
throughout the guided trajectory, proving the result.
\end{proof}
\end{Corollary}

\section{Laplacian-Constrained Stiefel Flow Matching}
We now describe the implementation of our proposed \textit{Laplacian-Constrained Stiefel Flow Matching} (LaSFM) model, including the architecture, hyperparameters and training procedure.

\subsection{Model} \label{app:model}

\begin{figure}[t]
    \centering
    \includegraphics[width=\linewidth]{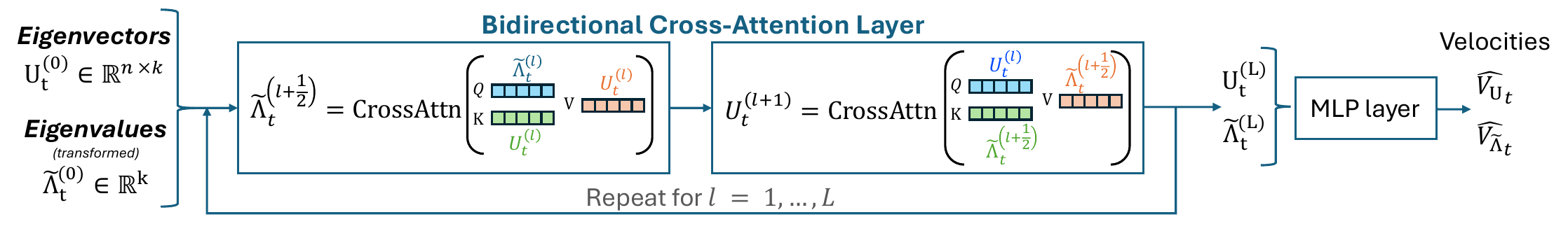}
    \caption{Overview of LaSFM architecture.}
    \label{fig:lasfm_architecture}
\end{figure}
\paragraph{Architecture.}
Following the coupled spectral architecture of
GGSD~\citep{minello2025generating}, LaSFM jointly processes the eigenvector
matrix $U_t\in\mathbb{R}^{n\times k}$ and the transformed eigenvalues
$\widetilde{\Lambda}_t\in\mathbb{R}^k$. Each row of $U_t$ is represented as an
eigenvector token, while each entry of $\widetilde{\Lambda}_t$ is represented
as an eigenvalue token. Both token streams are conditioned on a sinusoidal
embedding of the flow time
$t$~\citep{vaswani2017attention,ho2020denoising}. The eigenvalue tokens also
receive learned spectral-index embeddings, while graph size is encoded using
the fraction of valid nodes.

The two token streams interact through stacked bidirectional cross-attention
blocks. In each block, the eigenvalue tokens are first updated using the
eigenvector tokens, after which the eigenvector tokens are updated using the
resulting eigenvalue tokens. This allows information to be exchanged between
the eigenvalue and eigenvector representations throughout the network.

The resulting eigenvalue tokens at time $t$ predict the transformed-eigenvalue
velocity $\widehat{V}_{\widetilde{\Lambda}}$. We use this velocity to estimate
the terminal transformed eigenvalues:
\[
\widehat{\widetilde{\Lambda}}_1
=
\widetilde{\Lambda}_t
+
(1-t)\widehat{V}_{\widetilde{\Lambda}}
\bigl(t,U_t,\widetilde{\Lambda}_t\bigr).
\]
An embedding of $\smash{\widehat{\widetilde{\Lambda}}_1}$ then provides an
additional update to the eigenvector tokens, allowing the eigenvector velocity
to depend on the predicted terminal spectrum. A linear head predicts an
ambient eigenvector velocity, which is projected onto the constrained Stiefel
tangent space.

This design allows the eigenvalues and eigenvectors to evolve jointly and
enables guidance to act on both components throughout generation. In contrast,
SFMG first completes eigenvalue generation and then conditions eigenvector
generation on the resulting spectrum~\citep{huang2025graph}.

\paragraph{Dataset standardization.}
For each dataset, we standardize the transformed eigenvalues using the training-set mean and standard deviation. After generation, we undo this standardization before applying the cumulative-softplus mapping to obtain the Laplacian eigenvalues.

\paragraph{Hyperparameters.} The dataset-specific hyperparameters used to train LaSFM are reported in Table~\ref{tab:lasfm_hyperparameters}.

\begin{table*}[t]
    \centering
    \caption{Hyperparameters used to train LaSFM on each dataset.}
    \label{tab:lasfm_hyperparameters}
    \small
    \setlength{\tabcolsep}{9pt}
    \renewcommand{\arraystretch}{1.10}
    \begin{tabular}{lcccc}
        \hline
        Hyperparameter
        & Planar
        & SBM
        & Community-small
        & QM9 \\
        \hline

        \multicolumn{5}{l}{\textit{Spectral representation}} \\
        Retained eigenpairs $k$
            & 16 & 4 & 8 & 8 \\
        \hline

        \multicolumn{5}{l}{\textit{Spectral Transformer}} \\
        Hidden dimension
            & 128 & 128 & 96 & 128 \\
        Number of layers
            & 4 & 8 & 3 & 4 \\
        Attention heads
            & 4 & 4 & 4 & 4 \\
        Time embedding dimension
            & 16 & 16 & 8 & 8 \\
        Dropout
            & 0.05 & 0.005 & 0.0 & 0.0 \\
        \hline

        \multicolumn{5}{l}{\textit{Training}} \\
        Batch size
            & 128 & 128 & 128 & 128 \\
        Training epochs
            & 20{,}000 & 10{,}000 & 10{,}000 & 6{,}000 \\
        Eigenvalue pretraining epochs
            & 1{,}000 & 800 & 1{,}000 & 500 \\
        Learning rate
            & $10^{-3}$ & $10^{-3}$ & $10^{-3}$ & $10^{-3}$ \\
        \hline

        \multicolumn{5}{l}{\textit{Learning-rate scheduler}} \\
        Reduction factor
            & 0.5 & 0.5 & 0.5 & 0.5 \\
        Patience
            & 250 & 250 & 400 & 250 \\
        Improvement threshold
            & 0.001 & 0.005 & 0.005 & 0.005 \\
        Minimum learning rate
            & $10^{-5}$ & $10^{-5}$ & $10^{-5}$ & $10^{-5}$ \\
        \hline
    \end{tabular}
\end{table*}

\subsection{Training}\label{app:training}
\paragraph{Sampling $U_0$ and $U_1$.}
For each data graph, we discard the constant eigenvector associated with the zero eigenvalue and retain the $k$ nonzero-frequency eigenvectors as $U_1$. Since $U_1$ is obtained from the eigendecomposition of a valid graph Laplacian, it already satisfies
\[
U_1^\top U_1=I_k,
\qquad
U_1^\top\mathbf{1}=0.
\]

To construct a base sample $U_0$ satisfying the same constraints, we draw a Gaussian matrix $Z\in\mathbb{R}^{n\times k}$. Then, we project it onto the subspace orthogonal to the constant vector $\allone$ using Eq.~\eqref{eq:project_centered}, as follows:
\[
\Pi(Z) = PZ,
\]
where $P =  I_n - \frac{1}{n}\mathbf{1}\mathbf{1}^\top$. We then compute the QR decomposition of $\Pi(Z)$, i.e.,
\[
\Pi(Z) = QR,
\]
where the factor $Q$ is on the Stiefel manifold~\citep{tropp2012comparison}

Then we set $U_0=Q$. Since $Q$ has orthonormal columns that span a subspace of $\allone_\perp$, $U_0$ satisfies both
\[
U_0^\top U_0=I_k
\qquad\text{and}\qquad
U_0^\top\mathbf{1}=0.
\]


\newpage
\section{Experimental Details}\label{app:experiment-details}

\subsection{Datasets}
We evaluate on three synthetic graph datasets and one real-world molecular
dataset commonly used for graph generative models. We use the same preprocessed
datasets and 80\%/20\% training--test splits following SPECTRE~\citep{martinkus2022spectre}.

\textbf{Planar}: A synthetic dataset introduced in ~\citet{martinkus2022spectre}, containing 200 planar graphs with 64 nodes each. Each graph is constructed by uniformly sampling points in the plane and computing their Delaunay triangulation.

\textbf{Stochastic Block Model (SBM)}: A synthetic dataset from ~\citet{martinkus2022spectre}, containing 200 stochastic block-model graphs. Each graph has between two and five communities, with each community containing between 20 and 40 nodes. Edges are sampled with probability 0.3 within communities and 0.05 between communities, producing graphs with clear modular structure.

\textbf{Community-small}: A synthetic dataset containing 100 graphs with 12--20 nodes. Each graph consists of two approximately equal-sized and densely connected communities, with sparse edges connecting nodes from different communities.

\textbf{QM9}: A real-world molecular dataset containing approximately 134,000 small organic molecules~\citep{ramakrishnan2014quantum,wu2018moleculenet}. Each molecule contains at most nine heavy atoms drawn from carbon, nitrogen, oxygen, and fluorine. Molecules are represented as attributed graphs, where nodes denote atoms and edges denote chemical bonds.

\subsection{Evaluation Metrics}
\paragraph{Maximum Mean Discrepancy (MMD).}
We use the squared maximum mean discrepancy, $\mathrm{MMD}^2$~\citep{gretton2012kernel}, to assess the similarity between the generated and reference graph distributions. For two distributions $p$ and $q$,
\begin{equation}
\begin{aligned}
\mathrm{MMD}^2(p,q)
={}&
\mathbb{E}_{x,x'\sim p}\left[k(x,x')\right]
+
\mathbb{E}_{y,y'\sim q}\left[k(y,y')\right] \\
&-
2\mathbb{E}_{x\sim p,,y\sim q}\left[k(x,y)\right].
\end{aligned}
\label{eq:mmd}
\end{equation}
where $k$ is a positive-definite kernel. In practice, the expectations are estimated by averaging the kernel values over all pairs of empirical samples.

We use a Gaussian kernel over the total-variation distance,
\[
k(x,y)=\exp\left(
-\frac{d_{\mathrm{TV}}(x,y)^2}{2\sigma^2}
\right),
\quad
d_{\mathrm{TV}}(x,y)=\frac{1}{2}\lVert x-y\rVert_1,
\]
with \(\sigma=1\).

To account for the intrinsic train--test discrepancy caused by the finite dataset split, as well as differences arising from Laplacian normalization, we report the normalized MMD ratio
\[
r = \mathrm{MMD}^2(\text{generated}, \text{test})/\mathrm{MMD}^2(\text{train}, \text{test}),
\]
Lower values indicate that the generated distribution more closely matches the test distribution relative to the train--test baseline. We compute this ratio for the eigenvalue and wavelet statistics described below.

\paragraph{Eigenvalue MMD.}
For each graph, we sort the retained Laplacian eigenvalues in ascending order. In Eq.~\ref{eq:mmd}, \(p\) and \(q\) are the empirical distributions of these eigenvalue vectors for the generated and test graph sets (or train and test sets, respectively). This corresponds to the spectral MMD of~\citet{martinkus2022spectre}, except that we use the generated eigenvalues directly rather than reconstructing each graph and recomputing its eigendecomposition.

\paragraph{Wavelet MMD.}
Following~\citet{martinkus2022spectre}, we construct a multiscale descriptor
using 12 spectral wavelet filters~\citep{hammond2011wavelets}.
Each filter $g_s$ is a smooth band-pass function that emphasizes a different
range of Laplacian eigenvalues, allowing the descriptor to capture graph
structure at multiple frequency scales. For each filter and node $i$, we
compute
\[
S[s,i]
=
\sum_{\ell=1}^{k}
g_s(\lambda_\ell)^2 u_\ell[i]^2.
\]
The node responses for each
filter are summarized by a 100-bin histogram, and the 12 histograms are
concatenated into one wavelet descriptor. In Eq.~\ref{eq:mmd}, $p$ and $q$
denote the empirical distributions of these descriptors for the two graph sets
being compared. Unlike~\citet{martinkus2022spectre}, we compute the descriptors
directly from the generated eigenvalues and eigenvectors rather than from
reconstructed graphs.

\subsection{Training Details}

\paragraph{LaSFM training.}
We first pretrain the eigenvalue flow-matching model for the
number of epochs specified in Table~\ref{tab:lasfm_hyperparameters},
and then jointly train the eigenvalue and eigenvector models.
During pretraining, the eigenvalue model remains conditioned on
the intermediate eigenvector state $U_t$, obtained from the
prescribed Stiefel geodesic in Eq.~\ref{eq:geodesic_path}.
We select the checkpoint with the lowest training loss, smoothed
using a moving-average window of 100 epochs.

\paragraph{Baselines.}
We use the recommended hyperparameters and available
implementations for each baseline, while using the same data
split as LaSFM. For spectral evaluation, we use the same number
of retained eigenpairs $k$ across methods, since $k$ directly
affects the resulting MMD values. We follow the GGSD choice of
$k$ for each dataset, except for SBM: using $k=32$ was
computationally prohibitive, so we use $k=4$ for all methods,
consistent with the settings used by SPECTRE and SFMG. As the
SFMG implementation was not publicly available, we reproduce it
following \citet{huang2025graph}.

\paragraph{Computation.}
All experiments were conducted on a single NVIDIA GeForce
RTX 3090 GPU with 24\,GB of memory.

\subsection{Graph Generation}
\label{app:graph-generation}

\paragraph{Graph predictor.}
For the graph-generation experiment
(Sec.~\ref{sec:exp_graph}), we use a learned graph predictor to
map a generated spectral representation to a discrete graph.
Following \citet{huang2025graph}, the predictor takes
\[
\widehat L
=
\widehat U\widehat\Lambda\widehat U^\top
\]
as input and uses Euclidean flow matching to generate the
corresponding adjacency matrix.

The predictor is trained on Laplacian--adjacency pairs from the
training set. At generation time, we sample
$(\widehat U,\widehat\Lambda)$, construct $\widehat L$, and
apply the predictor to obtain an adjacency matrix, which is
then symmetrized and discretized. Training hyperparameters are
reported in Table~\ref{tab:fm_predictor_hyperparameters}.
We select checkpoints from the Pareto frontier of
generated-to-training MMD and VUN, balancing distributional
matching and graph quality without using the test set.

We use this learned reconstruction model only for the
graph-generation evaluation. For the property-guidance and
ablation experiments, we instead reconstruct graphs directly
from the generated Laplacian by thresholding, so that the
evaluation more directly reflects the quality of the generated
eigensystem.

\paragraph{Graph evaluation.}
Following GGSD, degree, clustering, and orbit statistics are
represented as normalized histograms before computing MMD with
a Gaussian total-variation kernel. We use bandwidths
$\sigma=1$ for degree, $\sigma=0.1$ for clustering, and
$\sigma=30$ for node-normalized 4-node orbit counts.
For Planar, a graph is valid if it is planar and connected.
For SBM, validity follows the standard benchmark criterion
based on recovered community structure, within- and
between-community connection probabilities, and community-size
constraints. Uniqueness is the fraction of valid generated
graphs that are pairwise non-isomorphic, novelty is the
fraction not isomorphic to any training graph, and VUN is the
fraction satisfying validity, uniqueness, and novelty
simultaneously.

\paragraph{Baselines.}
For spectral evaluation, we fix $k$ across methods because the
eigenvalue and wavelet descriptors are computed directly from the
retained eigensystem and therefore depend directly on $k$.
For graph-generation evaluation, the dependence on $k$ is indirect,
through the reconstructed graph. We therefore evaluate each method
using its recommended value of $k$, following the corresponding paper.

\begin{table}[t]
    \centering
    \caption{
    Hyperparameters used to train the flow-matching
    graph predictor for each dataset.
    }
    \label{tab:fm_predictor_hyperparameters}
    \small
    \setlength{\tabcolsep}{5pt}
    \renewcommand{\arraystretch}{1.05}
    \begin{tabular}{l|ccc}
        \hline
        Hyperparameter & Planar & Community-small & SBM \\
        \hline
        Batch size       & 8                & 16               & 16 \\
        Learning rate    & $2\times10^{-4}$ & $1\times10^{-3}$ & $5\times10^{-5}$ \\
        Hidden dimension & 256              & 512              & 256 \\
        Weight decay     & $1\times10^{-4}$ & $1\times10^{-3}$ & $5\times10^{-3}$ \\
        Number of blocks & 8                & 4                & 4 \\
        \hline
    \end{tabular}
\end{table}

\subsection{Property Guidance}

\paragraph{Background.}
Gradient-based guidance steers a generative process toward a
desired target by modifying its vector field using the gradient
of a target-dependent objective
\citep{dhariwal2021diffusion,holderrieth2025introduction}.
For generation conditioned on a target property $y^\star$,
Bayes' rule gives
\[
\nabla_x \log p_t(x\mid y^\star)
=
\nabla_x \log p_t(x)
+
\nabla_x \log p_t(y^\star\mid x),
\]
which motivates augmenting the generative dynamics with a
gradient that increases compatibility with the target.

For continuous properties, we use a pretrained spectral
predictor $f:(\Lambda,U,X)\mapsto y$ and define
\[
\mathcal{L}_{\mathrm{guide}}(\Lambda_t,U_t)
=
\left(f(\Lambda_t,U_t,X)-y^\star\right)^2.
\]
We therefore modify the generative vector field using the
negative gradient of $\mathcal{L}_{\mathrm{guide}}$, with a
guidance strength $\alpha$ controlling the magnitude of the
update. For LaSFM, this update is applied separately to the
eigenvalue and eigenvector components as described below.

\paragraph{Guidance predictor.}
For each target property, we train a separate spectral Transformer
$f:(\Lambda,U,X)\mapsto y$ that predicts the molecular property
from the Laplacian eigensystem and atom types $X$. The model
constructs node-level spectral features from $U$, $\Lambda U$,
$U^2$, and $\Lambda U^2$, concatenates them with atom features,
and processes them with a Transformer encoder. During training,
we randomly flip eigenvector signs to account for sign ambiguity
and optimize mean squared error on standardized property targets.
We select the checkpoint with the lowest validation MAE.

\paragraph{LaSFM guidance.}
We apply the guidance update separately to the transformed-eigenvalue and
eigenvector vector fields:
\begin{equation}
\begin{aligned}
V_{\widetilde{\Lambda}}^{\mathrm{guided}}
(t,\widetilde{\Lambda}_t,U_t)
&=
V_{\widetilde{\Lambda}}
(t,\widetilde{\Lambda}_t,U_t)
-
\alpha
\nabla_{\widetilde{\Lambda}_t}
\mathcal{L}_{\mathrm{guide}},\\
V_U^{\mathrm{guided}}
(t,\widetilde{\Lambda}_t,U_t)
&=
V_U
(t,\widetilde{\Lambda}_t,U_t)
-
\alpha
\widetilde{\Pi}_{U_t}
\left(
\nabla_{U_t}\mathcal{L}_{\mathrm{guide}}
\right),
\end{aligned}
\label{eq:guidance}
\end{equation}
where $\alpha$ controls the guidance strength. The eigenvalue
gradient is backpropagated through the mapping in
Eq.~\ref{eq:lambda_map}, while the eigenvector gradient is
projected using Eqs.~\ref{eq:project_centered}
and~\ref{eq:pproj}. This preserves both eigenvector
orthonormality and orthogonality to the constant vector during
guidance. We tune the guidance strength separately for each method and target property over \(\alpha \in \{0.1, 0.5, 1, 4, 8, 16, 32\}\), selecting the value that best balances target MAE and molecular validity.

\paragraph{Baseline guidance.}
We use the same property objective for all methods, adapting the
gradient update to their respective sampling procedures. For
GGSD~\citep{minello2025generating}, guidance is applied at each
reverse diffusion step. Since the DDPM noise predictor determines
the score of the noisy marginal up to a known scale, the
property-likelihood gradient can be incorporated as an additional
score term in the reverse denoising update
\citep{ho2020denoising,dhariwal2021diffusion}.

For SFMG~\citep{huang2024can}, guidance follows its two-stage
generation procedure. The eigenvalue flow is first guided toward
the target property, after which the eigenvector flow is guided
conditional on the generated eigenvalues. This preserves the
sampling structure of the original method while introducing the
same property objective used for LaSFM.

\subsubsection{Molecular Reconstruction and Evaluation}
We evaluate whether guidance in spectral space translates to the
desired properties after reconstructing the corresponding molecules.

\paragraph{Molecular reconstruction.}
Since we use full-rank generation ($k=8$ for nine-heavy-atom
QM9 molecules), we directly reconstruct the Laplacian and obtain
binary molecular connectivity by thresholding $-L_{ij}$. The
threshold is selected on unguided samples to maximize molecular
validity and is then fixed for guided generation. Atom types are
inherited from sampled QM9 molecules
\citep{ramakrishnan2014quantum}. Since the Laplacian specifies
connectivity but not bond order, we use RDKit~\citep{rdkit} to
assign single, double, or triple bonds from the generated
connectivity and atom types, followed by molecular sanitization.

\paragraph{Independent property evaluator.}
For each target property, we train a separate directed
message-passing neural network (D-MPNN)~\citep{yang2019analyzing}
to predict HOMO, LUMO, or the HOMO--LUMO gap from the
reconstructed molecular graph. Each evaluator takes the atom
types and RDKit-assigned bond types as input and is trained on
the corresponding QM9 target. These evaluators are trained
independently from the spectral predictors used for guidance
and are used only for evaluation.


\subsection{Ablation}

\paragraph{Metrics.}
We evaluate the ablations using reconstruction error
$\mathcal{E}_{\mathrm{recon}}$ and molecular validity.
Given a generated eigensystem $(\widehat U,\widehat\Lambda)$,
we reconstruct the spectral Laplacian as
$\widehat L=\widehat U\widehat\Lambda\widehat U^\top$
and threshold $\widehat L$ to obtain a discrete graph.
The reconstruction error is defined as
\[
\mathcal{E}_{\mathrm{recon}}
=\|\widehat L-L_{\mathrm{recon}}\|_F/
(\|\widehat L\|_F+\epsilon),
\]
where $L_{\mathrm{recon}}$ is the Laplacian of the reconstructed
graph. Lower values indicate greater consistency between the
generated spectral representation and the reconstructed graph.

We additionally report molecular validity. A molecule is considered
valid if its heavy-atom graph is connected and RDKit successfully
assigns bond orders and sanitizes the molecule.

\begin{table}[t]
    \centering
    \caption{
    Property-guided QM9 generation in spectral space.
    LaSFM achieves lower MAE with greater reduction compared to unguided generation.
    }
    \label{tab:guidance_spectral}
    \small
    \setlength{\tabcolsep}{2pt}
    \renewcommand{\arraystretch}{1.15}

    \begin{tabular}{@{}l|cc|cc|cc@{}}
        \hline
        & \multicolumn{2}{c|}{$\epsilon_{\mathrm{HOMO}}$}
        & \multicolumn{2}{c|}{$\epsilon_{\mathrm{LUMO}}$}
        & \multicolumn{2}{c}{$\Delta\epsilon$} \\
        \hline

        Method
        & MAE\,\downbetter & Gain\,\upbetter
        & MAE\,\downbetter & Gain\,\upbetter
        & MAE\,\downbetter & Gain\,\upbetter \\
        \hline

        GGSD
        & 1.3720 & 12\%
        & 0.9835 & 6.2\%
        & 1.9680 & 2.6\% \\

        SFMG
        & 0.1532 & 85\%
        & 0.4065 & 81\%
        & 0.3267 & 84\% \\

        LaSFM
        & \textbf{0.0277} & \textbf{97\%}
        & \textbf{0.0885} & \textbf{96\%}
        & \textbf{0.1046} & \textbf{95\%} \\

        \hline
    \end{tabular}
\end{table}

\begin{table*}[t]
    \centering
    \caption{Spectral constraint errors before and after gradient-based guidance on QM9 across molecular property targets (lower is better). LaSFM consistently ensures no error with and without guidance.}
    \label{tab:guidance_error}
    \small
    \setlength{\tabcolsep}{3.2pt}
    \renewcommand{\arraystretch}{1.15}
    \begin{tabular}{l|ccc|ccc|ccc}
        \hline
        & \multicolumn{3}{c|}{$\epsilon_{\mathrm{HOMO}}$}
        & \multicolumn{3}{c|}{$\epsilon_{\mathrm{LUMO}}$}
        & \multicolumn{3}{c}{$\Delta\epsilon$} \\
        Method
        & $\mathcal{E}_{\mathrm{orth}}$\,\downbetter
        & $\mathcal{E}_{\mathrm{null}}$\,\downbetter
        & $\mathcal{E}_{\lambda}$\,\downbetter
        & $\mathcal{E}_{\mathrm{orth}}$\,\downbetter
        & $\mathcal{E}_{\mathrm{null}}$\,\downbetter
        & $\mathcal{E}_{\lambda}$\,\downbetter
        & $\mathcal{E}_{\mathrm{orth}}$\,\downbetter
        & $\mathcal{E}_{\mathrm{null}}$\,\downbetter
        & $\mathcal{E}_{\lambda}$\,\downbetter \\
        \hline
        GGSD
        & 0.1080 & 0.0137 & 0.0003
        & 0.1080 & 0.0137 & 0.0003
        & 0.1080 & 0.0137 & 0.0003 \\
        \quad guided
        & \textcolor{red}{$\uparrow$\,0.1133}
        & \textcolor{red}{$\uparrow$\,0.0155}
        & 0.0003
        & 0.1062
        & 0.0124
        & 0.0003
        & \textcolor{red}{$\uparrow$\,0.1082}
        & \textcolor{red}{$\uparrow$\,0.0138}
        & 0.0003 \\
        \hline
        SFMG
        & 0.000 & 0.2114 & 0.000
        & 0.000 & 0.2114 & 0.000
        & 0.000 & 0.2114 & 0.000 \\
        \quad guided
        & 0.000
        & \textcolor{red}{$\uparrow$\,0.2444}
        & \textcolor{red}{$\uparrow$\,0.0423}
        & 0.000
        & \textcolor{red}{$\uparrow$\,0.2493}
        & \textcolor{red}{$\uparrow$\,0.0146}
        & 0.000
        & \textcolor{red}{$\uparrow$\,0.2501}
        & 0.000 \\
        \hline
        LaSFM
        & 0.000 & 0.000 & 0.000
        & 0.000 & 0.000 & 0.000
        & 0.000 & 0.000 & 0.000 \\
        \quad guided
        & \textbf{0.000} & \textbf{0.000}  & \textbf{0.000} 
        & \textbf{0.000}  & \textbf{0.000} & \textbf{0.000} 
        & \textbf{0.000}  & \textbf{0.000}  & \textbf{0.000}  \\
        \hline
    \end{tabular}
\end{table*}

\newpage
\section{Additional Experimental Results}
\label{app:additional-results}

\subsection{Additional Guidance Results}

\paragraph{Spectral guidance and constraint preservation.}
We first evaluate guidance directly in spectral space. We report
the mean absolute error (MAE) between the target property and the
value predicted from the generated eigensystem, together with the
relative MAE reduction,
\[
\left(
1-\frac{\mathrm{MAE}_{\mathrm{guided}}}
        {\mathrm{MAE}_{\mathrm{unguided}}}
\right)\times 100\%.
\]
To quantify whether guidance disrupts the spectral constraints,
we additionally measure
\begin{equation}
\label{eq:errors}
\mathcal{E}_{\mathrm{orth}}
= \frac{\|\hat U^\top \hat U-I_k\|_F}{\sqrt{k}},
\qquad
\mathcal{E}_{\mathrm{null}}
= \frac{\|\hat L\mathbf{1}/\sqrt{n}\|_2}
        {\|\hat L\|_F+\epsilon},
\qquad
\mathcal{E}_{\lambda}
= \frac{\|\operatorname{ReLU}(-\hat\lambda)\|_2}
        {\|\hat\lambda\|_2+\epsilon},
\end{equation}
where
$\hat L=\hat U\hat\Lambda\hat U^\top$.
These measure violations of eigenvector orthonormality, the
Laplacian nullspace constraint, and eigenvalue nonnegativity,
respectively.

As shown in Table~\ref{tab:guidance_spectral}, LaSFM achieves the lowest
MAE across all three molecular property targets and the largest
improvement over its unguided counterpart, indicating more
effective steering in spectral space.
Table~\ref{tab:guidance_error} further shows that LaSFM maintains
negligible constraint violations both before and after guidance.
In contrast, GGSD and SFMG exhibit larger constraint violations
in their unguided samples, which are further amplified by
guidance. These results show that LaSFM enables effective
property steering while preserving the spectral constraints.

\begin{figure}[h]
    \centering
    \includegraphics[width=0.5\linewidth]{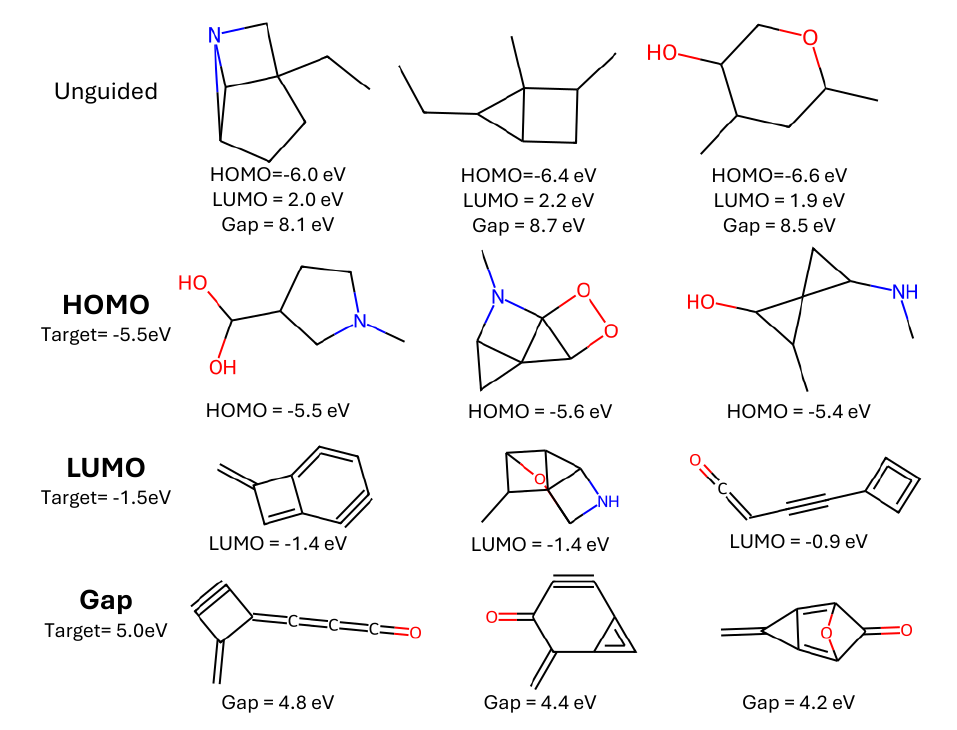}
    \caption{Reconstructed QM9 molecules from unguided and property-guided LaSFM. Guidance steers the generated molecules toward the target HOMO, LUMO, and HOMO--LUMO gap values.}
    \label{fig:guided_qm9_graphs}
\end{figure}

\paragraph{Guided molecule visualization.}
Figure~\ref{fig:guided_qm9_graphs} shows representative QM9
molecules reconstructed from unguided and guided LaSFM samples.
The guided samples qualitatively illustrate how spectral guidance
shifts the generated molecules toward the target HOMO, LUMO, and
HOMO--LUMO gap values while retaining chemically valid examples.

\subsection{Additional Ablation Results}
\paragraph{Spectral guidance and constraint-preservation ablation.}
We again report the target MAE, relative improvement over unguided
generation, and spectral constraint errors from
Eq.~\ref{eq:errors}. Table~\ref{tab:ablation-spectral} compares guidance
performance across the ablation settings. The full LaSFM model
achieves the lowest MAE and largest improvement for every target,
indicating that preserving all three constraints contributes to
effective property guidance.

\begin{figure}[h]
    \centering
    \includegraphics[width=0.5\linewidth]{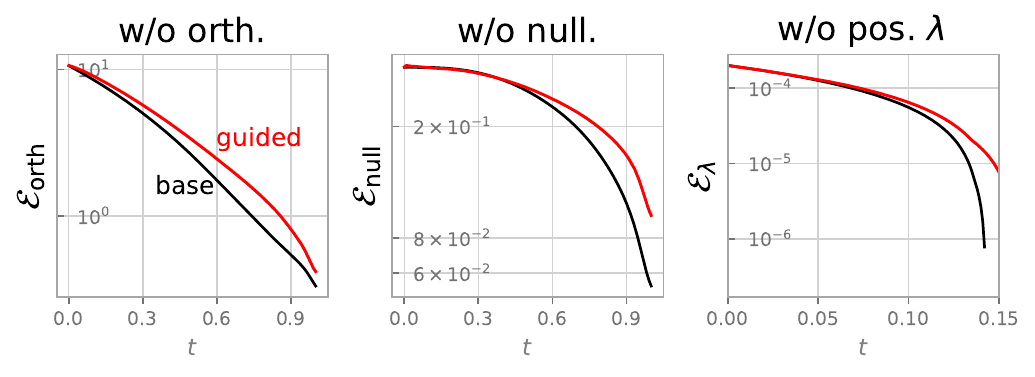}
    \caption{Ablation errors over generation process for $\epsilon_\mathrm{LUMO}$. Guidance increases constraint violations, although the eigenvalue-positivity error vanishes by $t=0.15$.}
    \label{fig:ablation_generation_error}
\end{figure}

Figure~\ref{fig:ablation_generation_error} shows that removing
any constraint increases the corresponding spectral violations,
which are further amplified during guidance. These deviations are
consistent with the higher MAE and smaller guidance gains observed
for the ablated models.

\begin{table}[h]
    \centering
    \caption{Guidance ablation on QM9. Removing any Laplacian constraint degrades performance (higher MAE and lower gain over unguided), especially orthonormality and the null-space constraint.}
    \label{tab:ablation-spectral}
    \small
    \setlength{\tabcolsep}{3pt}
    \renewcommand{\arraystretch}{1.1}
    \begin{tabular}{@{}l|cc|cc|cc@{}}
        \hline
        & \multicolumn{2}{c|}{$\epsilon_{\mathrm{HOMO}}$}
        & \multicolumn{2}{c|}{$\epsilon_{\mathrm{LUMO}}$}
        & \multicolumn{2}{c}{$\Delta\epsilon$} \\
        Method
        & MAE\,\downbetter & Gain\,\upbetter
        & MAE\,\downbetter & Gain\,\upbetter
        & MAE\,\downbetter & Gain\,\upbetter \\
        \hline
        LaSFM & \textbf{0.0277} & \textbf{97\%} & \textbf{0.0885} & \textbf{96\%} & \textbf{0.1046} & \textbf{95\%} \\
        \quad w/o orth.
        & 0.0849 & 91\% & 0.2654 & 87\% & 0.2486 & 87\% \\
        
        \quad w/o null.
        & 0.0872 &  91\% & 0.2270 & 89\% & 0.2675 & 86\% \\

        \quad w/o pos. $\lambda$
        & 0.0294 & 97\% & 0.1041 & 95\% & 0.1511 & 92\%\\
        \hline
    \end{tabular}
\end{table}

\end{document}